%% file: main.tex
\documentclass[letterpaper, 10 pt, journal, twoside]{ieeetran}

\usepackage{graphics} % for pdf, bitmapped graphics files
\usepackage{epsfig} % for postscript graphics files
\usepackage{mathptmx} % assumes new font selection scheme installed
\usepackage{times} % assumes new font selection scheme installed
\usepackage{amsmath} % assumes amsmath package installed
\usepackage{amssymb}  % assumes amsmath package installed
\usepackage{algorithm,algorithmic}
\usepackage{textcomp}
\usepackage{wrapfig}
\usepackage{subcaption}
\usepackage{multirow}
\usepackage{array}
\usepackage{hyperref}
\usepackage{indentfirst}
\usepackage{siunitx}
\usepackage[colorinlistoftodos]{todonotes}

\usepackage{bm}
\usepackage{float}

\newcommand{\RN}[1]{%
  \textup{\uppercase\expandafter{\romannumeral#1}}%
}
\newcommand{\etal}{\textit{et al. }}

\title{\LARGE \bf
{A Study of Shared-Control with Force Feedback for Obstacle Avoidance in Whole-body Telelocomotion of a Wheeled Humanoid}}

\author{DongHoon Baek$^{1\dagger}$, Yu-Chen (Johnny) Chang$^{2\dagger}$, and Joao Ramos$^{1,2}$% <-this % stops a space
\thanks{This work is supported by the National Science Foundation via grants IIS-2024775 and CMMI-2043339.}
\thanks{$\dagger$DongHoon Baek and Yu-Chen (Johnny) Chang contributed equally.}
\thanks{The authors are with the $^1$ Department of Mechanical Science and Engineering and the $^2$ Department of Electrical and Computer Engineering at the University of Illinois at Urbana-Champaign, USA. {\tt\small dbaek4@illinois.edu, yuchenc2@illinois.edu}} 
}

\begin{document}

\maketitle
%\thispagestyle{empty}
%\pagestyle{empty}
% Paper headers
% \markboth{IEEE Robotics and Automation Letters. Preprint Version. Accepted October, 2021}

%%%%%%%%%%%%%%%%%%%%%%%%%%%%%%%%%%%%%%%%%%%%%%%%%%%%%%%%%%%%%%%%%%%%%%%%%%%%%%%%
\begin{abstract}
Teleoperation has emerged as an alternative solution to fully-autonomous systems for achieving human-level capabilities on humanoids. Specifically, teleoperation with whole-body control is a promising hands-free strategy to command humanoids but demands more physical and mental effort. To mitigate this limitation, researchers have proposed shared-control methods incorporating robot decision-making to aid humans on low-level tasks, further reducing operation effort. However, shared-control methods for wheeled humanoid telelocomotion on a whole-body level has yet to be explored. In this work, we study how whole-body feedback affects the performance of different shared-control methods for obstacle avoidance in diverse environments. A Time-Derivative Sigmoid Function (TDSF) is proposed to generate more intuitive force feedback from obstacles. Comprehensive human experiments were conducted, and the results concluded that force feedback enhances the whole-body telelocomotion performance in unfamiliar environments but could reduce performance in familiar environments. Conveying the robot's intention through haptics showed further improvements since the operator can utilize the force feedback for short-distance planning and visual feedback for long-distance planning.
\end{abstract}
% Humanoid robots have emerged as a promising platform for diverse applications. However, how to develop a fully-autonomous and capable humanoid is still an open research question. On the other hand, whole-body teleoperation is a promising alternative to remotely control robots to achieve human-level capabilities. To mitigate the physical and mental demand of teleoperation, researchers proposed shared-control methods that incorporate robot decision making to aid humans on low-level tasks to reduce operation effort. But shared-control with haptic feedback for wheeled humanoid telelocomotion on a whole-body level has yet to be explored. In this work, we studied how different shared-control methods with bilateral feedback perform for obstacle avoidance through humanoid telelocomotion. Specifically, we explored how whole-body level bilateral feedback affects the performance of shared-control methods in diverse environments. The time-derivative Sigmoid function was proposed to generate more intuitive haptic feedback. We conducted and evaluated comprehensive human experiments and suggested appropriate shared control methods for different environments based on our results. \textcolor{red}{Prof: and what were the results? Explain briefly.}  

%%%%%%%%%%%%%%%%%%%%%%%%%%%%%%%%%%%%%%%%%%%%%%%%%%%%%%%%%%%%%%%%%%%%%%%%%%%%%%%%


\input{1_introduction}

\input{2_Method}

\input{3_experiment}

\input{4_results_discussion}
\input{5_conclusion}

%\addtolength{\textheight}{-12cm}   % This command serves to balance the column lengths
                                  % on the last page of the document manually. It shortens
                                  % the textheight of the last page by a suitable amount.
                                  % This command does not take effect until the next page
                                  % so it should come on the page before the last. Make
                                  % sure that you do not shorten the textheight too much.

%%%%%%%%%%%%%%%%%%%%%%%%%%%%%%%%%%%%%%%%%%%%%%%%%%%%%%%%%%%%%%%%%%%%%%%%%%%%%%%%

%%%%%%%%%%%%%%%%%%%%%%%%%%%%%%%%%%%%%%%%%%%%%%%%%%%%%%%%%%%%%%%%%%%%%%%%%%%%%%%%

%%%%%%%%%%%%%%%%%%%%%%%%%%%%%%%%%%%%%%%%%%%%%%%%%%%%%%%%%%%%%%%%%%%%%%%%%%%%%%%%
%\section*{APPENDIX}

%Appendixes should appear before the acknowledgment.

%\section*{ACKNOWLEDGMENT}

%The preferred spelling of the word �acknowledgment� in America is without an �e� after the �g�. Avoid the stilted expression, �One of us (R. B. G.) thanks . . .�  Instead, try �R. B. G. thanks�. Put sponsor acknowledgments in the unnumbered footnote on the first page.

%\begin{thebibliography}{99}
\bibliographystyle{IEEEtran}
\bibliography{main.bib}
%\end{thebibliography}

\end{document}

%% file: 1_introduction.tex
\section{INTRODUCTION}
\label{S:1}
Humanoid robots have been in the spotlight for a long time due to their promising potential to address problems in diverse scenarios from elderly care to disaster response \cite{wang2015hermes,purushottam2022hands,ramos2019dynamic}. Despite the recent advancements, developing fully autonomous humanoid robots capable of achieving human-level adaptation in navigating harsh terrains and executing physical tasks is still extremely challenging. 

In face of the challenges, teleoperation emerged as an alternative solution where semi-autonomous humanoid robots are remotely controlled by humans\cite{ramos2018humanoid}. By integrating human's ability to make adaptive decisions in complicated environments with robot's physical advantages in precision and repeatability, teleoperation has been widely used in many robotics applications such as humanoids \cite{abi2018humanoid}, surgical robots\cite{hwang2018flexible}, mobile robots\cite{gottardi2022shared}, and robot manipulators\cite{luo2022human}. 

To teleoperate humanoids effectively, many human-machine interfaces (HMI) were developed to deliver human commands and receive feedback from the robot \cite{abi2018humanoid,wang2021dynamic,ramos2015balance}. However, teleoperation is challenging to employ when (1) the mapping between the operator's HMI inputs and the robot's subsequent motion is not intuitive, (2) the objects and obstacles are located outside of the camera's field of field or partially occluded by the robot, or (3) the visual feedback becomes mentally burdening in complicated environments. Such conditions are likely to increase the operator's effort and degrade performance while putting the robot in dangerous situations.  

% Shared-control
Shared-control has emerged as a promising approach to alleviate these limitations \cite{abbink2012haptic} and can be categorized into various strategies \cite{udupa2021shared}. To offload the mental and physical demand during teleoperation, a portion of the control authority can be assigned to the robot through the shared-control framework by detecting the human's intention and providing robot's feedback to the human \cite{losey2018review}. Gottardi \etal suggested a shared-control framework that utilized artificial potential fields (APF) to compensate the controller input by adding virtual repulsive and attractive points for better robot navigation \cite{gottardi2022shared}. Wang \etal proposed an adaptive servo-level shared-control scheme that combined human control input and obstacle avoidance controllers to assist people with disability through a mobile robot \cite{wang2014adaptive}. Song \etal used human confidence level gains decided by obstacle distance, operating speed, and operation time to decide the autonomous level in their shared-control framework for navigation \cite{song2016interactive}. Moreover, haptic feedback has been one of the effective approaches in shared-control applications for providing feedback to the human on the robot's state and the environment. Luo and Lin \etal designed a shared-control framework with haptic feedback that detects human intention through EMG signals for obstacle avoidance \cite{luo2019teleoperation}. Selvaggio \etal proposed a task-prioritized shared-control method using haptic guidance to inform the operator on kinematic constraints of a redundant manipulator \cite{selvaggio2019passive}. Rahal \etal presented a haptic-enabled shared-control framework to minimize the user's workload during a teleoperated manipulation task \cite{rahal2020caring}.  

\begin{figure*}[t]
\centerline{\includegraphics[width=15.5cm]{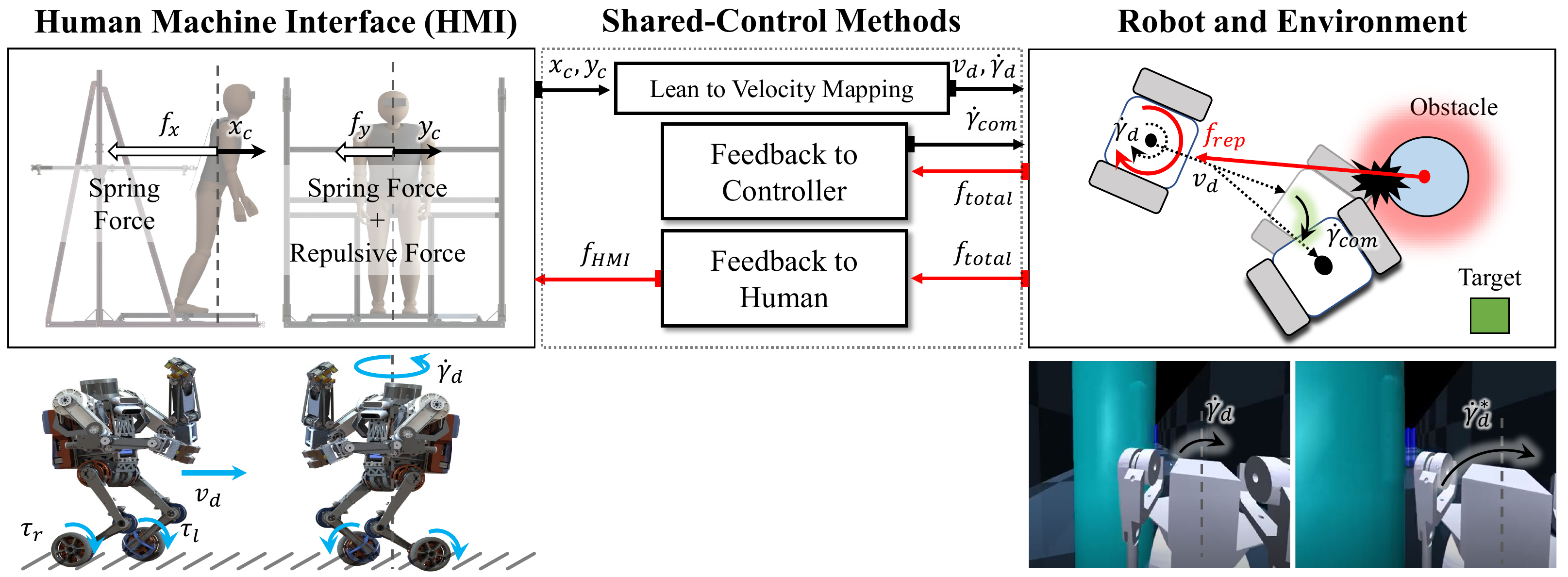}}
    \caption{\textbf{Shared-Control Framework.} The operator controls SATYRR through the whole-body HMI by moving the CoM. SATYRR moves forward and yaws when the operator leans in the sagittal and frontal planes. Two kinds of force feedback, F-H and F-C, aid the operator in avoiding collision by providing haptic feedback to the user and updating the velocity command. The obstacle repulsive force is calculated based on Eqn. \ref{time_derivative_sigmoid},\ref{sigmoid}, and \ref{force_total}.}
    \label{fig1}
\end{figure*}

However, \textit{no prior work has explored the effectiveness of providing force feedback through a whole-body HMI in a shared-control framework.} Such HMI with whole-body control and feedback allows hands-free telelocomotion, a subcategory of teleoperation where the operator remotely controls the legged robot's locomotion \cite{huang2020telelocomotion}. This allows the operator to navigate and perform manipulation tasks simultaneously on a humanoid robot for more dynamic motions \cite{purushottam2022hands} but requires more mental and physical effort compared to performing the tasks sequentially. Incorporating shared-control strategies could reduce such telelocomotion difficulty. Nevertheless, it remains to be seen if force feedback in a shared-control framework reduces the effort to avoid obstacles with whole-body telelocomotion on a humanoid robot. Additionally, the human preference for the level of autonomy and feedback methods in different environments require further investigation.

In this work, we explore and compare the performance of multiple shared-control approaches with force feedback for humanoid robot obstacle avoidance through whole-body telelocomotion. Our main contributions are highlighted as follows: (1) a novel shared-control framework with force feedback using whole-body telelocomotion with a Time-Derivative Sigmoid Function (TDSF); (2) an integration of the wheeled humanoid robot SATYRR in simulation, a whole-body HMI, and virtual reality (VR) equipment; (3) two stages of extensive human experiments to compare the performance and user preference of four different shared-control feedback methods.

%% file: 2_Method.tex
\section{METHOD and MATERIALS}
\label{method}
In this section, we introduce the different parts of the robotics system integrated in this work and describe how the obstacle repulsive force feedback is calculated through a Time-Derivative Sigmoid Function (TDSF). The methods utilized to apply the force feedback are described in details.
\subsection{Wheeled humanoid robot and Human-Machine Interface}
\label{system_description}

To evaluate different force feedback methods in our shared-control framework, a wheeled humanoid robot SATYRR \cite{purushottam2022hands} and a whole-body HMI \cite{wang2021dynamic} were utilized as described in Fig. \ref{fig1}. SATYRR is a wheeled bipedal robot consisting of a torso, two three degree-of-freedom (3DOF) arms, two legs with knee and ankle joints, and two wheels for traversal. As regards to the whole-body HMI, two back-drivable linear sensors and actuators (LISA) were used to detect the operator's center of mass (CoM) position while providing haptic force to the person's torso.

\subsection{Force Feedback Shared-Control Framework}
\label{Feedback Shared-Control Framework}

\subsubsection{Controller Design and Basic Teleoperation law}
The overview of our shared-control framework is shown in Fig. \ref{fig1}. The operator's CoM displacements along the $x$-axis and $y$-axis are mapped to the forward velocity $v_{\textrm{d}}$ and the angular velocity $\dot{\gamma}_{\textrm{d}}$ around the vertical axis of the robot respectively. Both of the desired velocities are fed into a Linear Quadratic Regulator (LQR) controller for balancing and a PD controller for turning, resulting with robot wheel torques ($\tau_\textrm{r}$ and $\tau_\textrm{l}$). We adopted a velocity mapping strategy \cite{purushottam2022hands} that uses different slopes and a dead-band to ensure any undesired small displacements in the operator's CoM does not result in high frequency movement for the robot. The wheeled inverted pendulum (WIP) model is utilized to stabilize the robot.

\subsubsection{Obstacle Repulsive Force from Time-Derivative Sigmoid-Function}
\begin{figure}[t] 
    \centering
	\includegraphics[width=0.9\columnwidth]{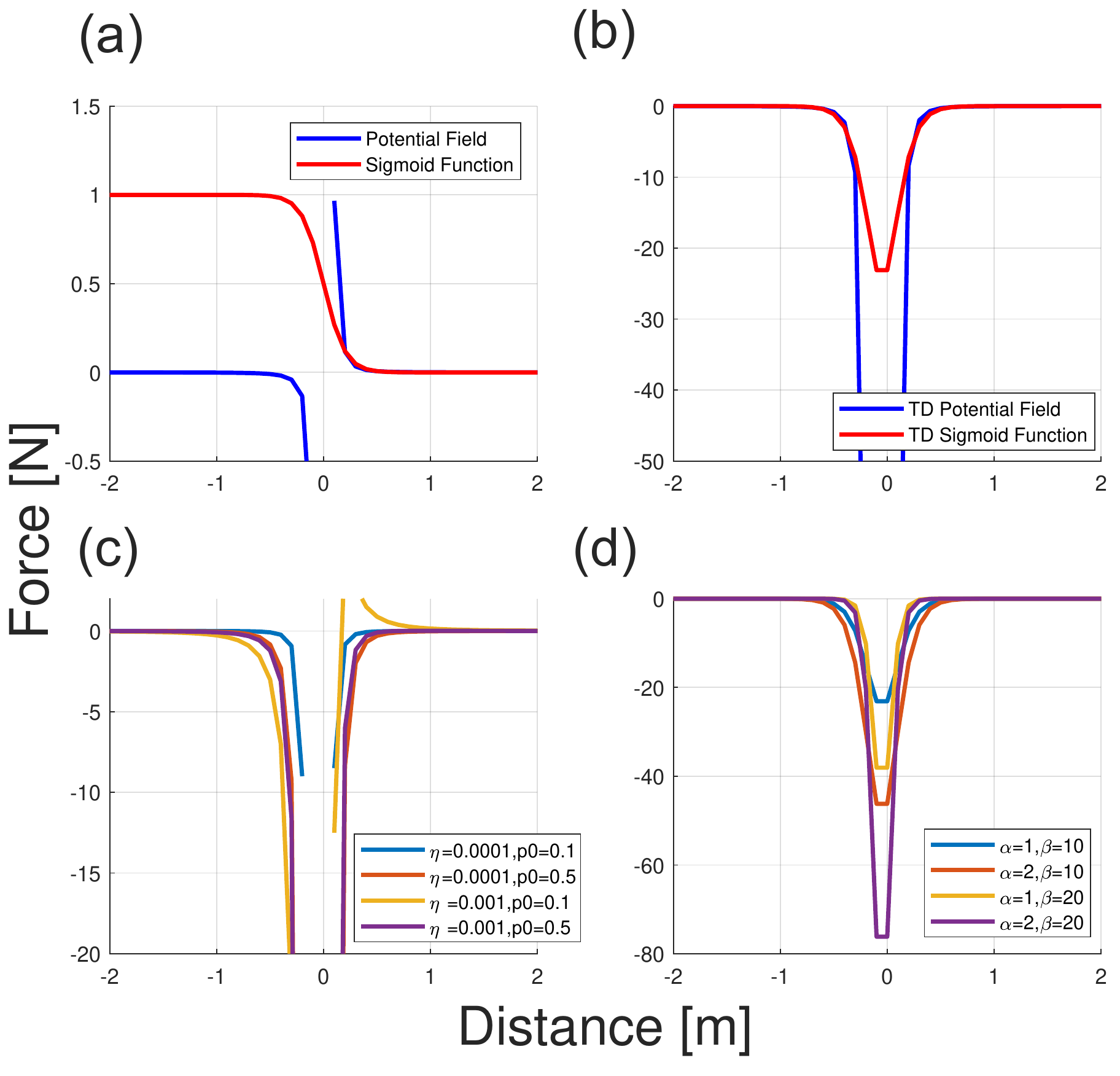}
        \caption{\textbf{Comparison of Force Profile.} All figures assume the same constant velocity $\frac{dp}{dt}$. Figures (a,b) indicate the force profiles for the time-derivative (TD) of APF and TDSF respectively. Figures (c,d) represent the resulting force profile of TD APF and TDSF respectively with various parameter values.}
	\label{force_comparison}
\end{figure}

To improve the operator's obstacle avoidance capabilities during telelocomotion, we generated repulsive feedback force for each obstacle to guide the robot away from potential collisions. Such repulsive force generated from the APF is commonly used in shared-control frameworks due to its low-computational cost \cite{luo2019teleoperation}.
%The repulsive force $f_{rep}$ of the obstacles \cite{khatib1986real} is calculated from the gradient of the potential given by:
% \begin{equation}\label{eqn_potential_field}
%     f_{\textrm{rep}}(p) \!\!=\!\!
%     \begin{cases}
%     \frac{\eta}{p^2}(\frac{1}{p}-\frac{1}{p_0})\nabla p &  p	\leq  p_0 \\
%        0  & p >  p_0
%     \end{cases}
% \end{equation}
% where $p$ represents the Euclidean distance between the robot and the obstacle while $p0$ represents the activation boundary of the force. The symbol $\eta$ is the hyperparameter for the magnitude of the force and $\nabla$ indicates the gradient.
Despite the APF's benefits, the force generated has several drawbacks when applied as whole-body feedback to the operator. Our preliminary tests repeatedly indicated that the operators felt uncomfortable being pushed or pulled by the obstacle repulsive force applied in the $x$-axis since they have less control over their balance in the sagittal plane while standing. The tests also showed that the operators felt discomfort from the constant feedback force encountered when cornering or slowly avoiding obstacles, considering that the magnitude of the APF force is solely determined by the Euclidean distances between the robot and the obstacles. Moreover, tuning the desired force profile of the APF is not intuitive since both of its hyperparameters (see \cite{1087247}) are coupled with the slopes of the force curves and activation distance as shown in Fig. \ref{force_comparison}. 

\begin{figure}[h]
     \centering
 \centering
 \includegraphics[width=1\linewidth]{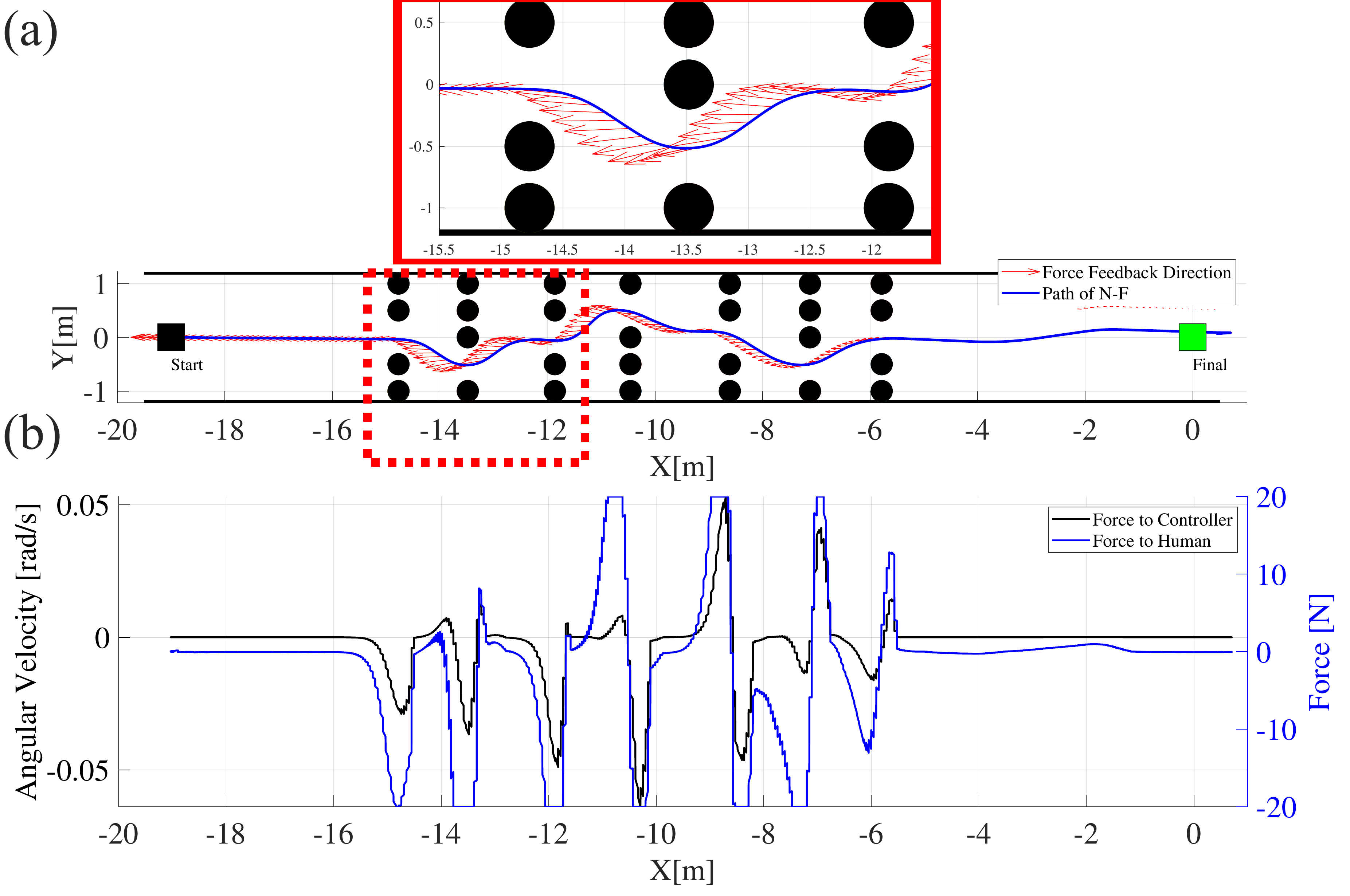}
 \caption{\textbf{Robot Trajectory and Force Feedback}. (a) The blue line is the N-F path, black circles are obstacles, and red arrows indicate the generated feedback force direction. (b) The graph shows how the force feedback is applied to the controller and the human.}
 \label{force_path}
\end{figure}

In order to generate a more intuitive feedback force that also considers the relative velocities between the robot and the obstacles, the Time-Derivative of a Sigmoid Function (TDSF) is proposed in place of the APF:

\begin{equation}\label{time_derivative_sigmoid}
    f_{\textrm{rep}}(p) \!\!=\!\!
    \begin{cases} 
    \frac{d}{dt}F_{\textrm{rep}}(p) = \beta\frac{dF_{\textrm{rep}}(p)}{dp}\frac{dp}{dt},    & \text{if }  p	\leq  p_0  \text{ and } \frac{dF_{\textrm{rep}}(p)}{dp} > 0\\
       0, & \text{otherwise}
  \end{cases}
\end{equation}
\begin{equation}\label{sigmoid}
    F_{\textrm{rep}}(p) = \frac{\alpha}{1+e^{\beta(-p)}}
\end{equation}

\noindent where $p$ is the distance to each obstacle, $p_0$ is the force activation distance, $\alpha$ is the force magnitude gain, and $\beta$ represents the gain for changing the slopes of the force curves. As shown in Fig. \ref{force_comparison}, the TDSF generates less aggressive force compared to the APF while the activation distance and the slopes of the force curve $\frac{dF_{\textrm{rep}}(p)}{dp}$ are easier to tune with $\alpha$ decoupled from the activation distance. The time-derivative term $\frac{dp}{dt}$ denotes that more force is applied when the relative velocity between the robot and the obstacle is large while no force is generated at static or constant relative velocity. Moreover, the derivative term $\frac{dF_{\textrm{rep}}(p)}{dp}$ is zero if the robot stays equidistant from the obstacle, generating no feedback force when the robot is not approaching potential collision. When $\frac{dF_{\textrm{rep}}(p)}{dp}$ is non-positive, $f_{\textrm{rep}}(p)$ is set to zero since the robot is distancing from the obstacle.
The total obstacle repulsive force $f_{\textrm{total}}$ is:
\begin{equation}\label{force_total}
    f_{\textrm{total}}(\bm{p},g(\bm{\theta})) = -(w_1\sum_{m=1}^{M}{f_{\textrm{rep}}(p_\textrm{m})g(\theta_\textrm{m})} + w_2\sum_{n=1}^{N}{f_{\textrm{rep}}(p_\textrm{n})g(\theta_\textrm{n})})     
\end{equation}

where $\bm{p}$ is the vector set of distance $p$ ($p_i \in \bm{p}$). Symbols $w_1$ and $w_2$ are the weights of the force from the obstacles and the walls while $M$ and $N$ denote the number of obstacles and walls respectively. The function $g(\theta)$ ($\theta_i \in \bm{\theta}$) takes in the angle between the obstacle and the robot's $x$-axis in the body frame and is defined differently in the next two sections for the feedback cases. An example of the force generated from the obstacles and the walls is illustrated in Fig. \ref{force_path}.

\subsubsection{Force Feedback to Controller}
One intuitive way for updating the robot's controller command input to consider obstacle avoidance is by adding the obstacle repulsive force $\dot\gamma_{\textrm{com}}$ to the operator's commanded velocity $\dot\gamma_{\textrm{d}}$ \cite{gottardi2022shared}, resulting in the updated angular velocity command $\dot\gamma^*_{\textrm{d}}$:
\begin{equation}\label{compensated_controller}
    \dot\gamma^*_{\textrm{d}} = \lambda \dot\gamma_{\textrm{d}} + \dot\gamma_{\textrm{com}}=\lambda \dot\gamma_{\textrm{d}} +f_{\textrm{total}}(\bm{p},g(\bm{\theta}))
\end{equation}
where $\lambda$ is parameter of sensitivity and $g(\theta)= \textrm{atan2}(\Vec{or}^{\textrm{i}}_{\textrm{y}},\Vec{or}^{\textrm{i}}_{\textrm{x}})$ represents the rotation angle of the yaw controller. Symbols $\Vec{or}^{\textrm{i}}_{\textrm{x}}$ and $\Vec{or}^{\textrm{i}}_{\textrm{y}}$ represent the vectors from the robot to each obstacle $i$ in the body frame's x-axis and y-axis. Only the yaw velocity command is compensated based on an important insight from preliminary tests that the operators disfavor velocity modification in the $x$-axis.

\subsubsection{Force Feedback to Human}
The total repulsive force $ f_{\textrm{total}}(\bm{p},g(\bm{\theta}))$ with $g(\theta) = \textrm{sin}(\textrm{atan2}(\Vec{or}^{\textrm{i}}_{\textrm{y}},\Vec{or}^{\textrm{i}}_{\textrm{x}}))$ is provided to the operator with a spring force that helps maintain the neutral stance position:
\begin{equation}\label{haptic_force_x}
    f_{\textrm{HMI}_\textrm{x}} = -k_{\textrm{x}}(x_{\textrm{H}}-x_{\textrm{H}_{\textrm{0}}})     
\end{equation}
\begin{equation}\label{haptic_force_y}
    % \begin{gathered}
    f_{\textrm{HMI}_\textrm{y}} = -k_\textrm{y}(y_\textrm{H}-y_{\textrm{H}_\textrm{0}}) - \mu f_{\textrm{total}}(\bm{p},g(\bm{\theta}))     
    % \end{gathered}
\end{equation}
where $k_\textrm{x}$ and $k_\textrm{y}$ are the spring constants. Symbol $\mu$ represents the customizable force feedback gain. Symbols $x_\textrm{H}$ and $y_\textrm{H}$ are the operator's CoM position while $x_{\textrm{H}_\textrm{0}}$ and $y_{\textrm{H}_\textrm{0}}$ are the calibrated neutral CoM position in the x-axis and y-axis. Two LISAs equally contribute to generating the forces $f_{\textrm{HMI}_\textrm{x}}$ and $f_{\textrm{HMI}_\textrm{y}}$ in the transverse plane. Note that the haptic force feedback is only applied in the $y$-axis while the spring force is applied in both $x$-axis and $y$-axis.

% \begin{algorithm}[h]
%   \caption{Shared-Control ($Combo$)}\label{algo:blah}
%   \begin{algorithmic}[1]
% %   \show\LOOP
%     \REQUIRE {Initialize the location of $M$ Obstacles randomly.} 
%     \ENSURE {Compensated torque $\tau_c$} 
%     \STATE \textbf{for} $n=0,...$ epoch \textbf{do} % Outer loop
%     \begin{ALC@g}
%     \STATE Select a single agent randomly among $M$ policies
%     \STATE \textbf{for} $m=0,...,M$ multiple agents \textbf{do}
%     \INDSTATE  Observe state $s_m$ and act an action $a_m$$\sim \pi_m(\cdot|s_m)$
%     \STATE \textbf{end for}
%     \STATE Compute a single univariate Gaussian distribution $\Pi(a|s) \sim \mathcal{N}(\mu_{\Pi},\,\sigma^{2}_{\Pi})$ (see Equation \ref{eqn_gmm1} and \ref{eqn_gmm2})
%     \STATE Compute the composite Gaussian distribution $\phi(a|s) \sim \mathcal{N}(\mu_{\phi},\,\sigma^{2}_{\phi})$  (see Equation \ref{eqn_composite})
    
%     \STATE $a = H(a|s) + \tanh{(\phi(a|s))}$
    
%     \STATE Execute $a$ and Observe next state $s^{'}$, reward $r$, and done signal $d$
%     \STATE Store $(s,a,r,s^{'},d)$ in replay buffer $\mathcal{D}$
%     \STATE If $s'$ is terminal, reset environment state.
%     \STATE \textbf{If} it's time to update \textbf{then}
%     \begin{ALC@g}
%     \STATE Compute targets and Update Q-function, policy, and target networks based on SAC algorithm \cite{haarnoja2018soft}
%     \end{ALC@g}
%     \textbf{end if}
%     \end{ALC@g}
%   \STATE \textbf{until} convergence
%   \end{algorithmic}
% \end{algorithm}

%% file: 3_experiment.tex
\section{EXPERIMENT}
\label{experimet}
To exhaustively compare the effectiveness of the shared-control methods under diverse conditions, we conducted two stages of human experiments with static and dynamic obstacles under different brightness conditions and feedback cases. Concretely, the first stage involved known maps with unvarying obstacle locations and velocities that were familiar to the operators while the second stage involved unknown maps with randomized obstacle configurations. The stages were designed based on our hypothesis that the operators will be less dependent on the robot's assistance in known environments where the optimal paths do not deviate while more reliant on the robot's decision making and the additional repulsive force feedback in unknown or mentally taxing environments. The details of each map can be seen in Fig. \ref{fig_map_design}. 

\subsection{Participants}
Five subjects completed both stages of experiments while two more subjects completed the stage two experiments. More subjects were recruited for the stage two experiments considering the randomness of the obstacle maps. All subjects were males within an age range of 22-34. Two of the five subjects had prior experience with teleoperating robots. All subjects have experience with driving cars. The experiments were approved and conducted in compliance with the requirements from the Internal Review Board (IRB) of UIUC.

\subsection{Experiment Setup}
\begin{figure}[t]
 \centering
 \includegraphics[width=1\linewidth]{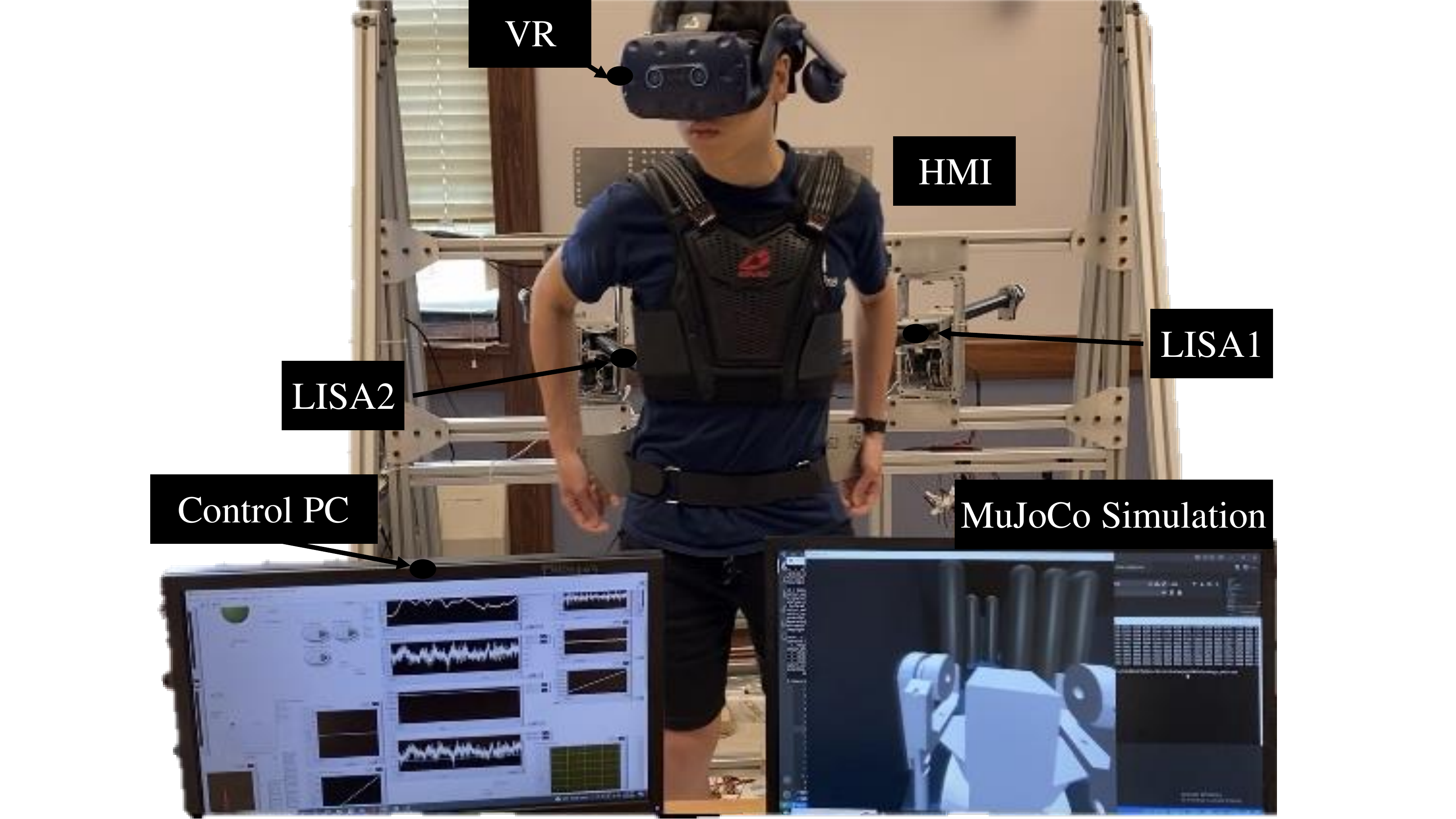}
 \caption{\textbf{Experiment Setup}.}
 \label{EXP_setting}
\end{figure}

To accurately calculate the feedback force for each obstacle and wall in the maps, we integrated a SATYRR model in the MuJoCo physics simulation such that the robot can obtain the global positions of all obstacles within its force feedback activation distance. In all experiments, the operators were equipped with a VR headset (VIVE Pro Eye, HTC, Taiwan) that provided the view of a virtual camera attached to the robot as shown in Fig. \ref{EXP_setting}. To reduce possible nausea from VR, the virtual camera was attached above and behind the robot such that the operator can see the robot's shoulder at all time. A low pass filter was implemented for the camera's yaw that tracks the robot's yaw, mimicking a camera stabilizer that eliminates high frequency movement. The User Datagram Protocol (UDP) was used to communicate between the HMI and the MuJoCo simulation through a wired Ethernet connection with a five to ten millisecond time-delay.

\begin{figure}[t] 
    \centering
	\begin{subfigure}{0.9\linewidth}
 		\includegraphics[width=\columnwidth]{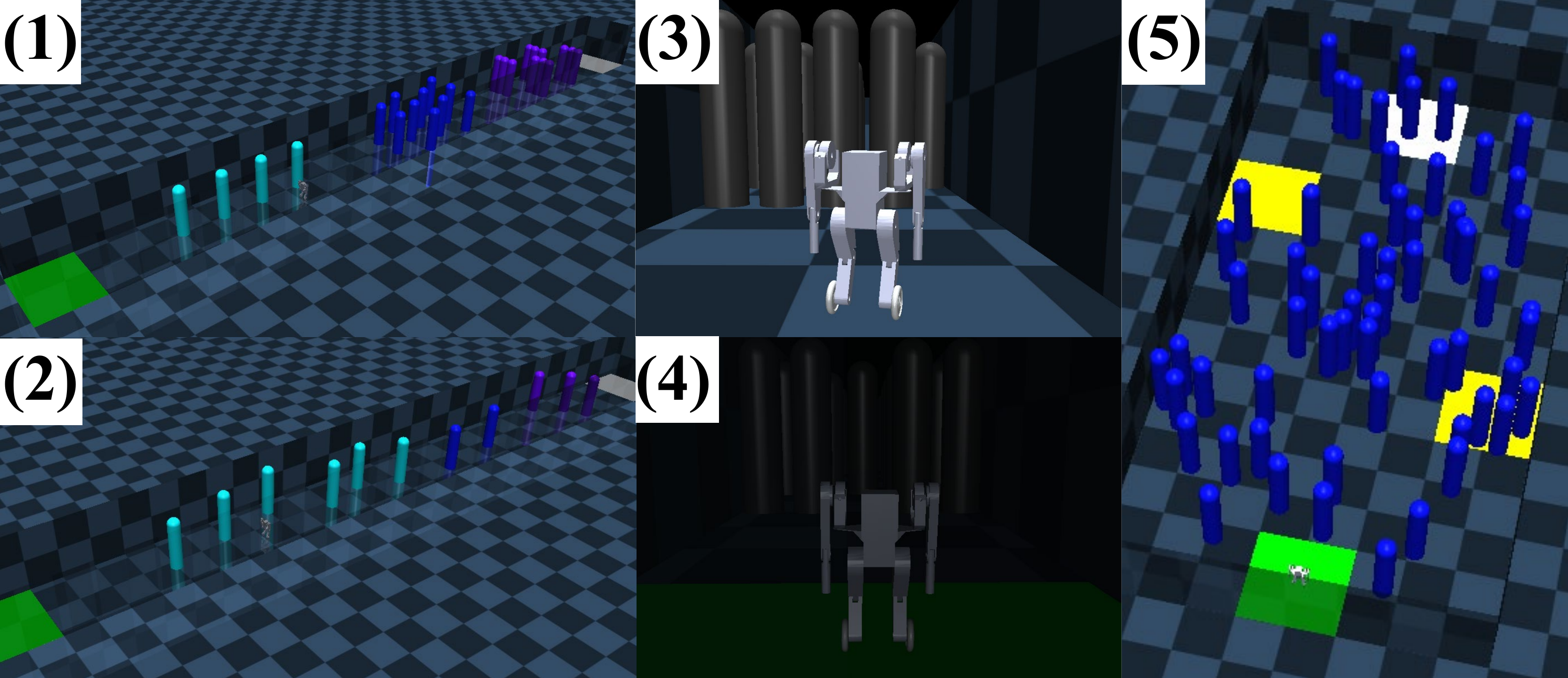}
 		\caption{Maps in MuJoCo Simulation}
 	\end{subfigure} %\hspace{0.05cm}
  
 	\begin{subfigure}{0.9\linewidth}
 		\includegraphics[width=\columnwidth]{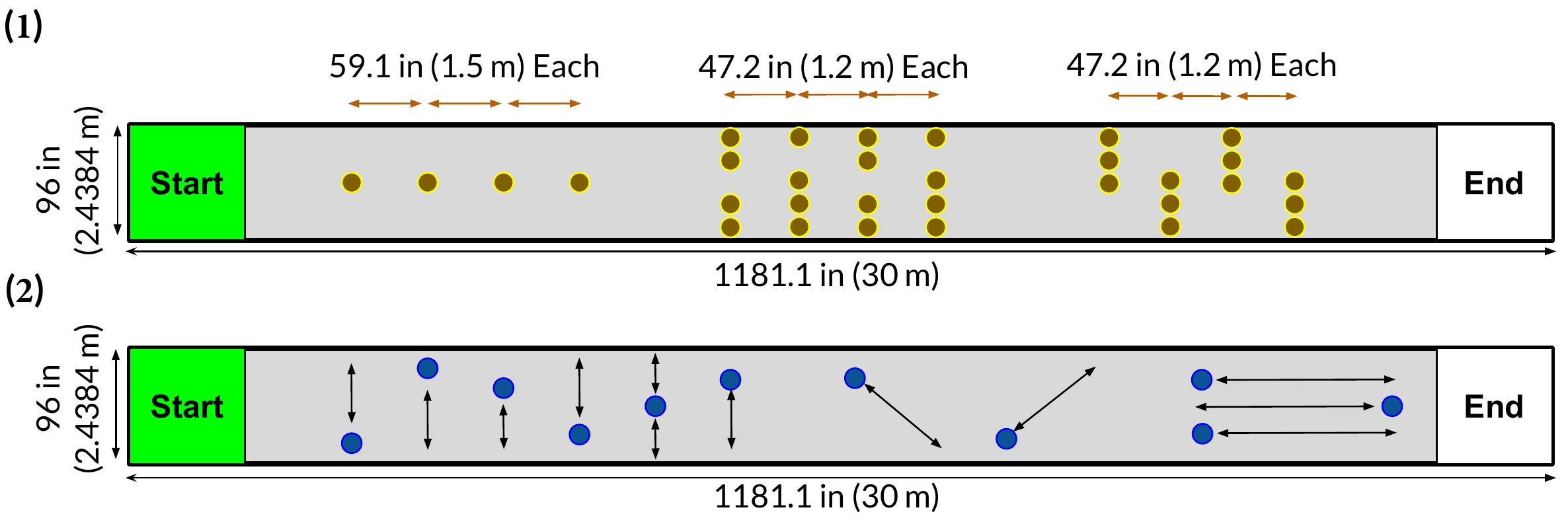}
 		\caption{Stage One Map Design}
 	\end{subfigure} %\hspace{0.05cm}

        \begin{subfigure}{0.9\linewidth}
 		\includegraphics[width=\columnwidth]{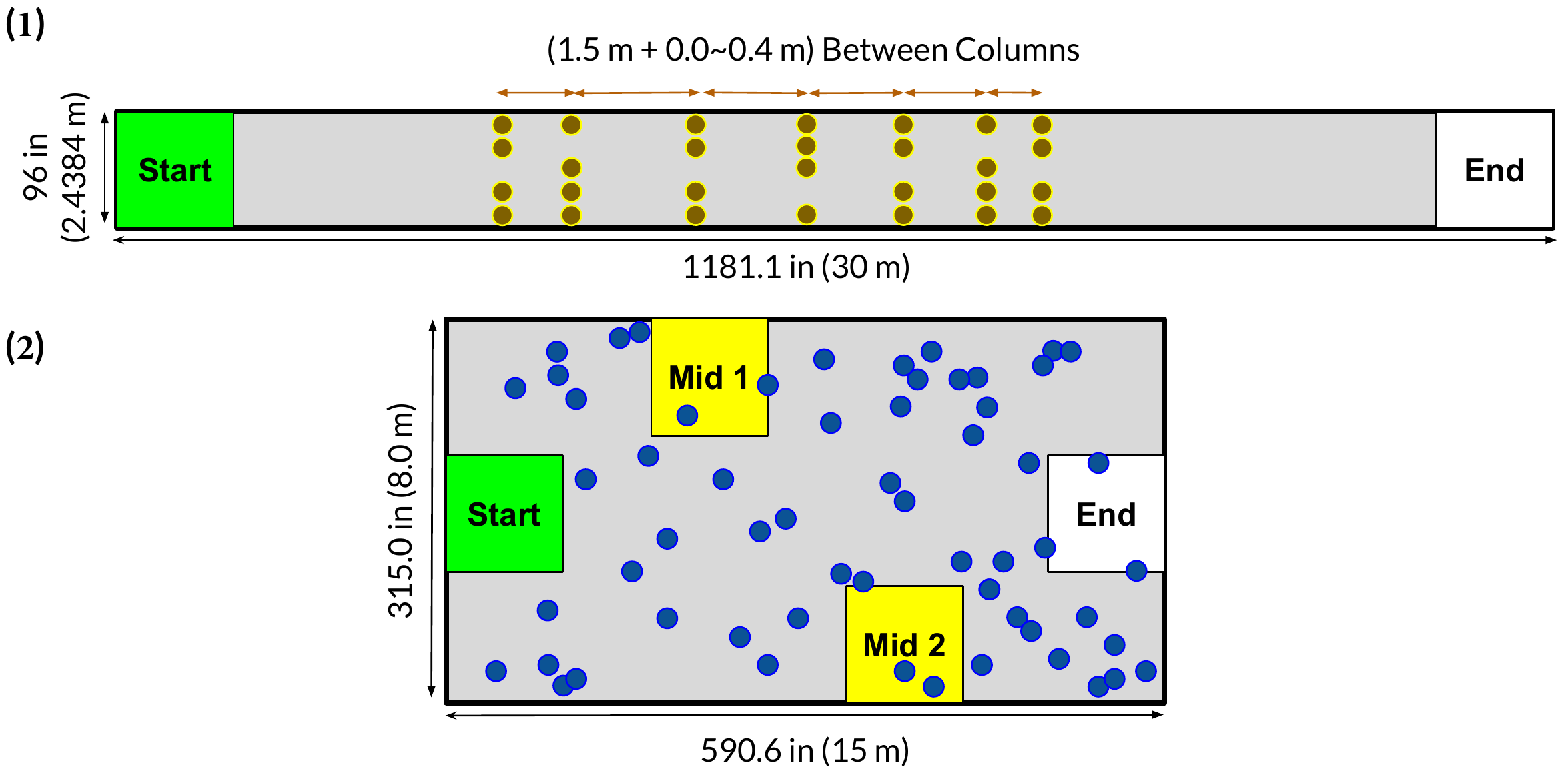}
 		\caption{Stage Two Map Design}
 	\end{subfigure} %\hspace{0.05cm}
    \caption{\textbf{Map Design}. (a.1) Stage One Known Static Map, (a.2) Stage One Known Dynamic Map, (a.3) Stage Two Unknown Bright Static Map, (a.4) Stage Two Unknown Dark Static Map, (a.5) Stage Two Unknown Dynamic Map, (b.1) Static Map, (b.2) Dynamic Map, (c.1) Bright and Dark Static Maps, and (c.2) Dynamic Map.}
    \label{fig_map_design}
\end{figure}

\subsection{Two Stage Experiment Map Design}
\subsubsection{Stage One with Known Maps}
The stage one experiments explored the influence of force feedback in environments familiar to the operator where the locations and velocities of all obstacles were predetermined. As shown in Fig. \ref{fig_map_design}, the experiments were conducted in two maps, one with static obstacles and one with dynamic obstacles. All obstacles have the same size and cylindrical shape, modeled from a human of 5'7'' with 7.9'' radius. In the static map, the operator has to pass the first four obstacles through a S-shaped route. The velocities of the dynamic obstacles are predetermined between 0.6m/s to 0.9m/s with fixed initial locations. 

% Static map:
% 35 obstacles, forming 7 columns
% Dark (10\% of the bright map with light source from the robot) vs bright maps
% Each column of 5 obstacles has one random obstacle that is absent
% Randomized distance between each column (varied turning angle and acceleration control)
% 5 trials for each case in both the dark and static static map

% Dynamic map:
% Dynamic map feedback tells you the shortest escape route: new map should be more spacious
% 2 randomized midpoints (left/right)
% 60 obstacles
% Random location and velocity
% Bounces off the walls
% 10 trials for each case in the dynamic map
\subsubsection{Stage Two with Unknown Maps}
The stage two experiments were designed to evaluate the effect of force feedback on the performance in unfamiliar and complex environments. The initial locations and velocities of the obstacles in the stage two static and dynamic maps were randomized in every experiment to prevent the subjects from learning the optimal paths. In the static map as shown in Fig. \ref{fig_map_design}, the distance between the columns of obstacles and the removed obstacle in each column are randomized to necessitate varied turning and acceleration control. The dynamic map, intentionally created to be very challenging, has 60 obstacles with two randomized required mid-points that appear close to either the top or bottom walls. In addition, we introduced a new dark static map that has 10\% brightness of the bright static map to examine sub-optimal visual feedback and its effect on the force feedback usage. In total, three stage two maps (static bright, static dark, and dynamic) were designed. 

\subsection{Shared-Control Feedback Cases}
Four shared-control feedback cases were tested in both stages of experiments: \\
(1) \textbf{No Feedback (N-F)}: basic telelocomotion without using any obstacle repulsive force. This is the method applied in the previous work \cite{purushottam2022hands}. \\
(2) \textbf{Feedback to Human (F-H)}: the obstacle repulsive force is applied as haptic feedback $f_{\textrm{HMI}}$ to the operator's torso. \\
(3) \textbf{Feedback to Controller (F-C)}: the obstacle repulsive force modifies the controller's reference angular velocity $\dot\gamma_{\textrm{d}}$ so the robot automatically adjusts its trajectory. \\
(4) \textbf{Combo}: a combination of F-H and F-C allows the operator to feel the haptic force feedback from F-H while F-C autonomously adjusts the robot's trajectory simultaneously. The force feedback activation distance for F-H is 25\% further than F-C to prevent the operator from feeling the haptic feedback lagging behind the automatic compensated trajectory.

\begin{figure*}[t]
     \centering
 	\begin{subfigure}[b]{0.325\linewidth}
 		\includegraphics[width=\columnwidth]{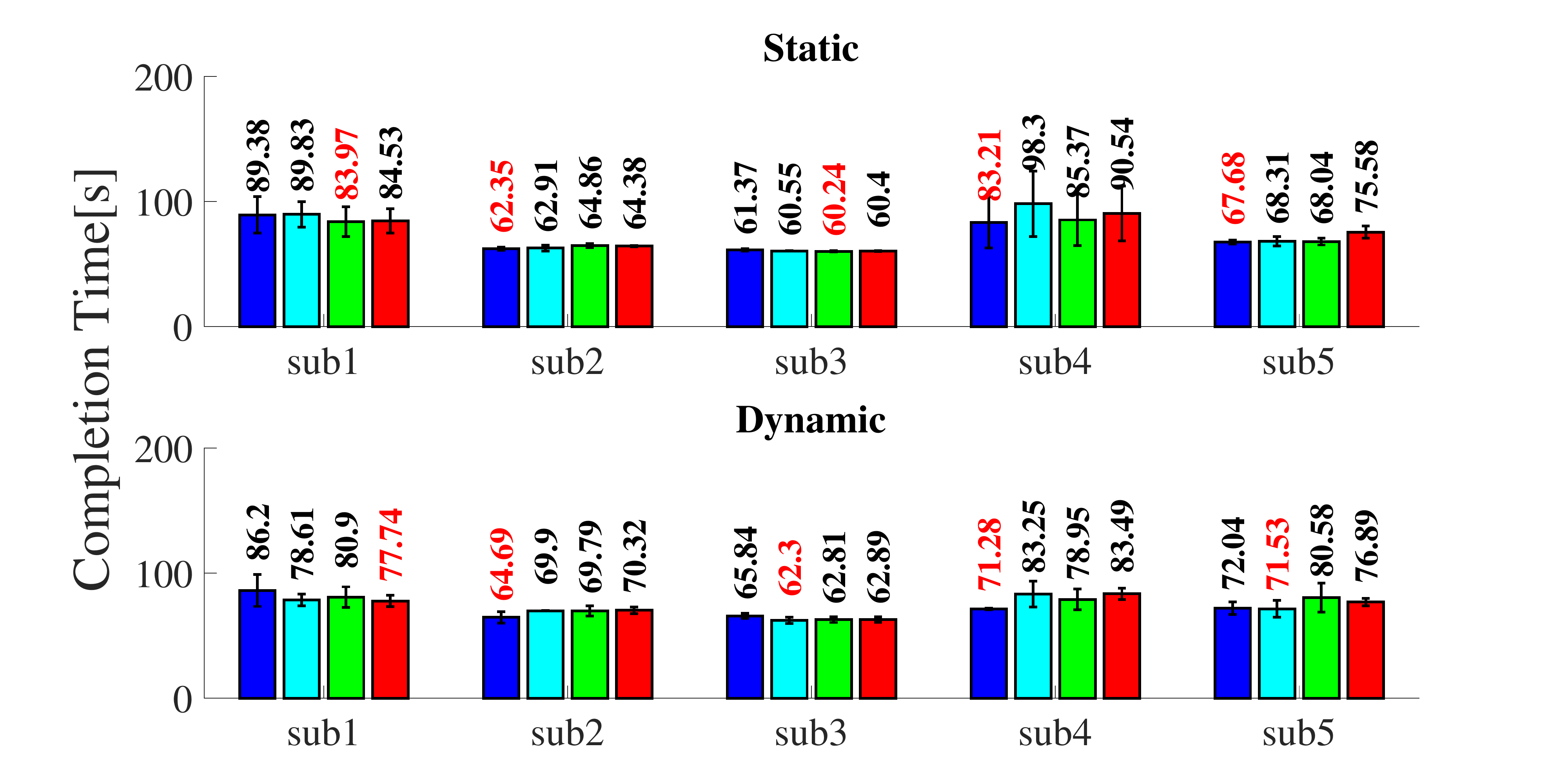}
 		\caption{}
 	\end{subfigure} %\hspace{0.05cm}
 	\begin{subfigure}[b]{0.325\linewidth}
 		\includegraphics[width=\columnwidth]{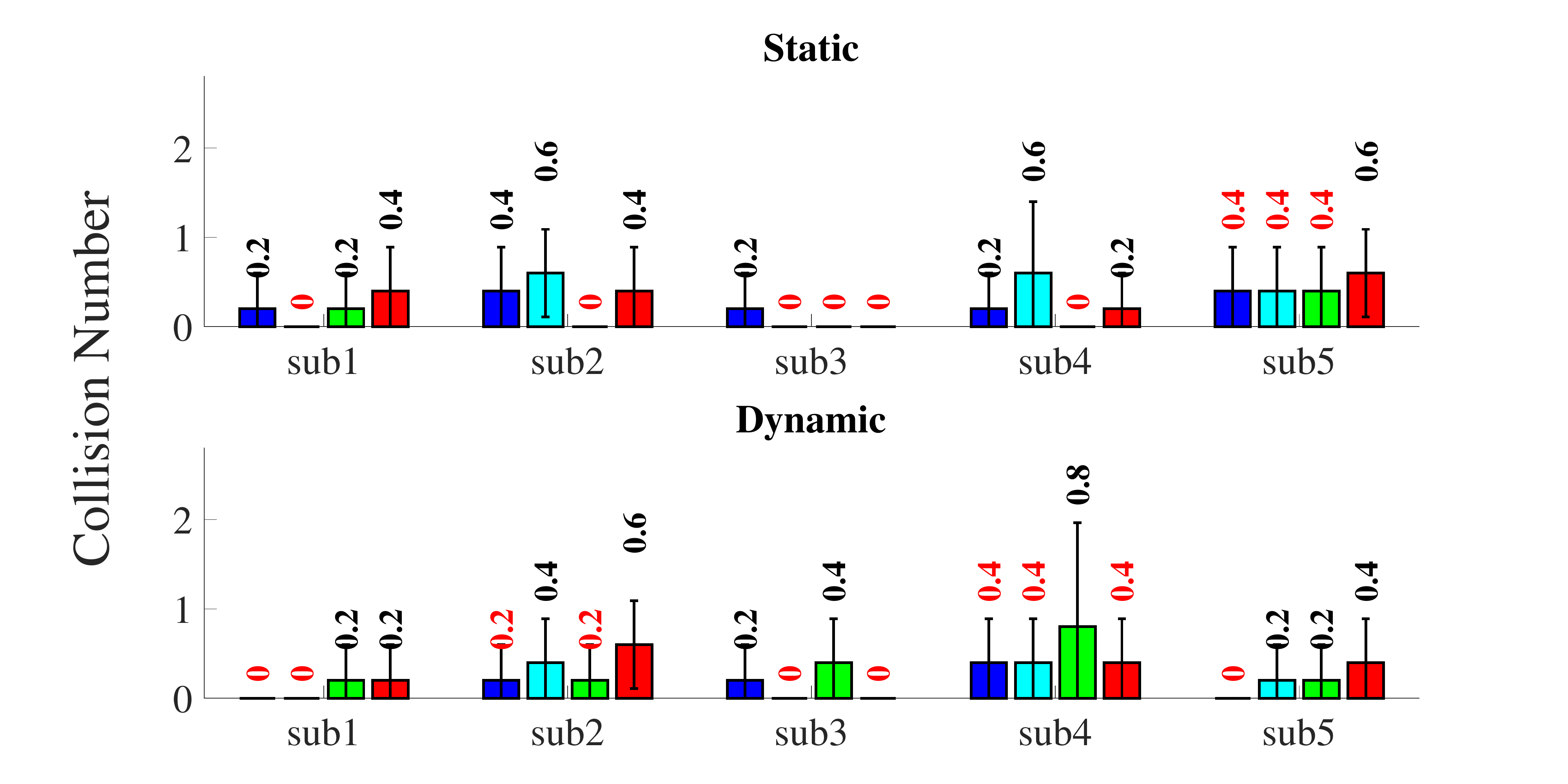}
 		\caption{}
 	\end{subfigure} %\hspace{0.05cm}
        \begin{subfigure}[b]{0.33\linewidth}
 		\includegraphics[width=\columnwidth]{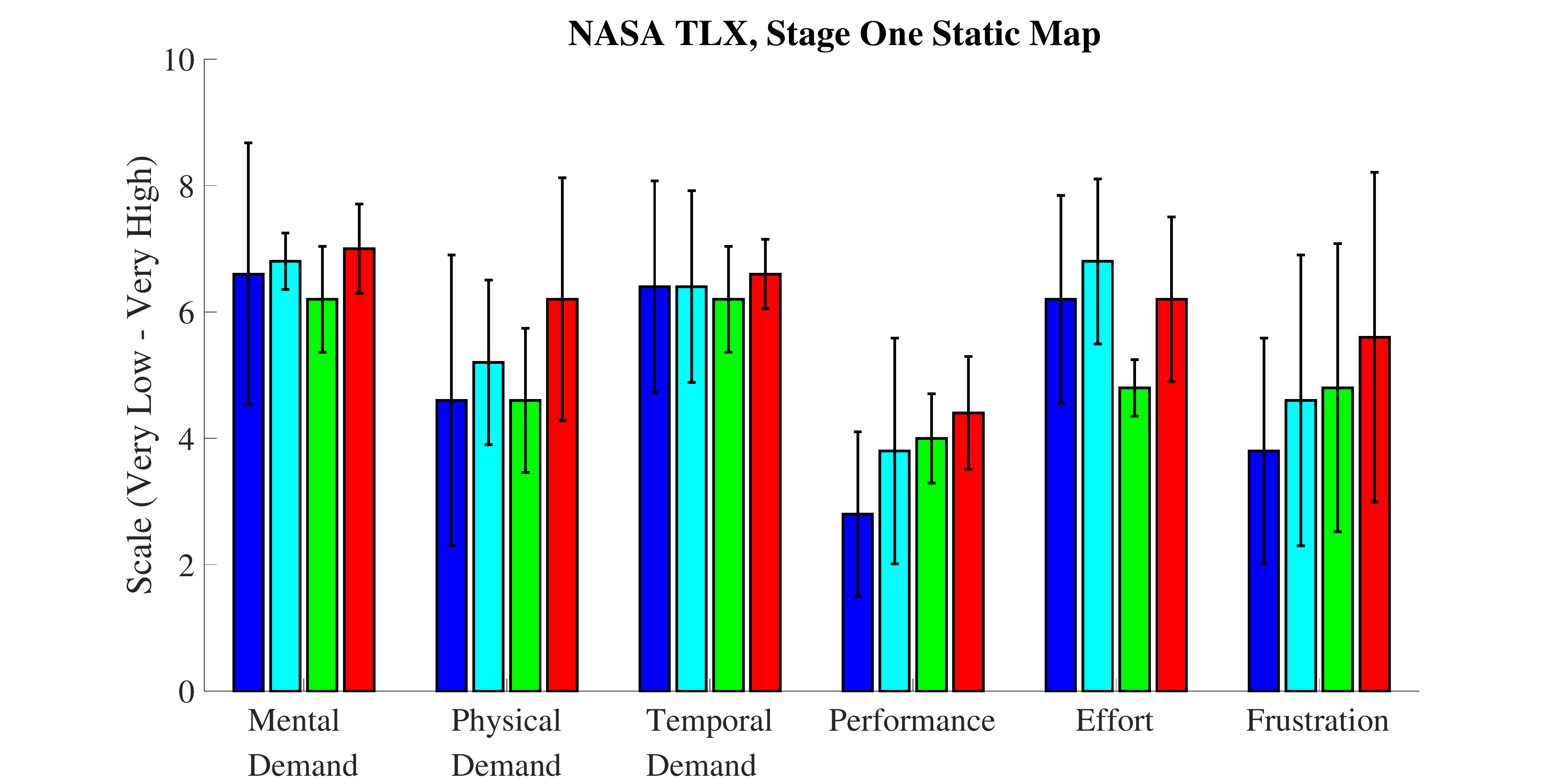}
 		\caption{}
 	\end{subfigure} %\hspace{0.05cm}
 	
 	\begin{subfigure}[b]{0.325\linewidth}
  		\includegraphics[width=\columnwidth]{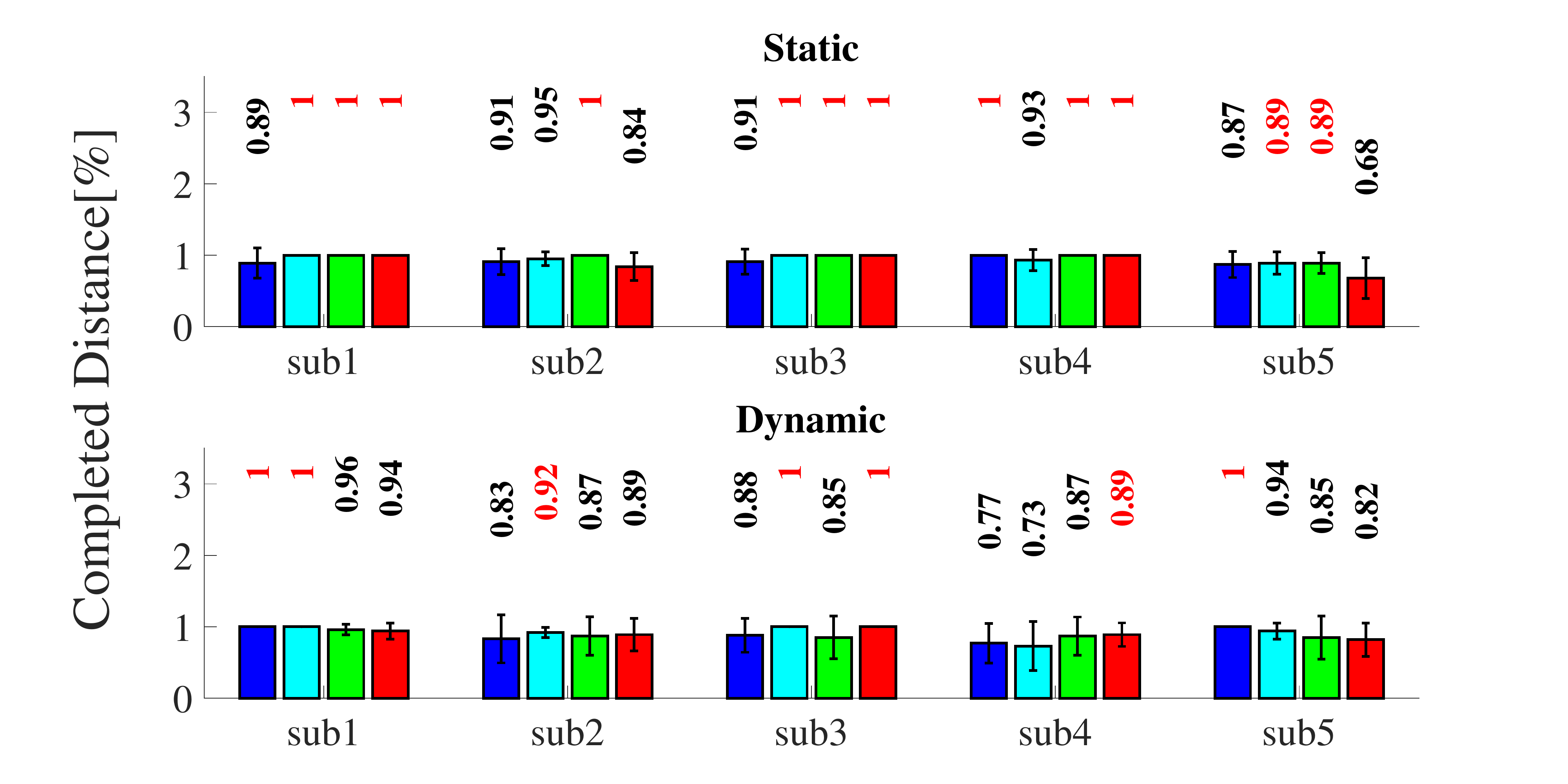}
 		\caption{}
 	\end{subfigure}
 	\begin{subfigure}[b]{0.325\linewidth}
  		\includegraphics[width=\columnwidth]{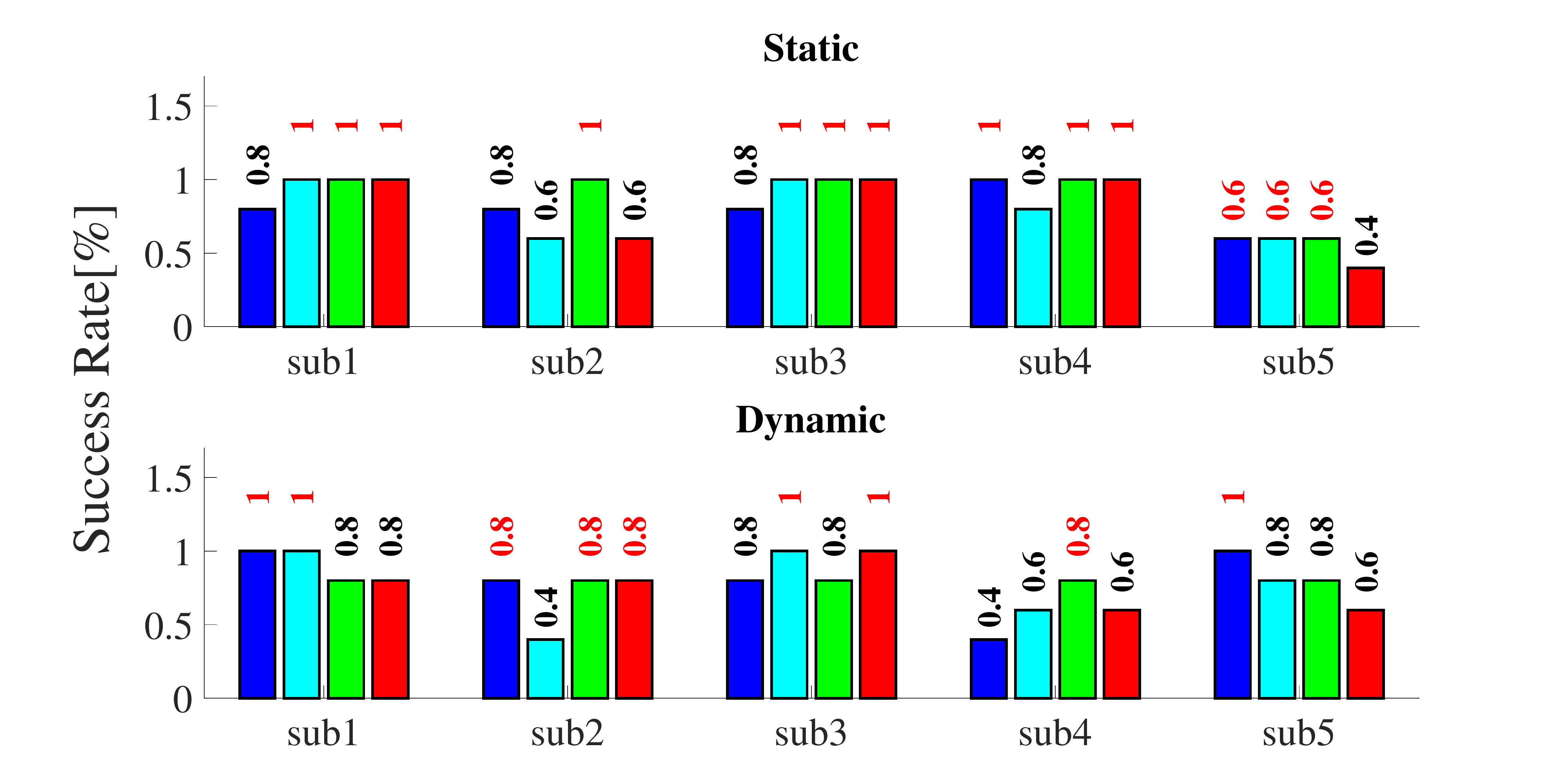}
 		\caption{}
 	\end{subfigure}
 	\begin{subfigure}[b]{0.325\linewidth}
 		\includegraphics[width=\columnwidth]{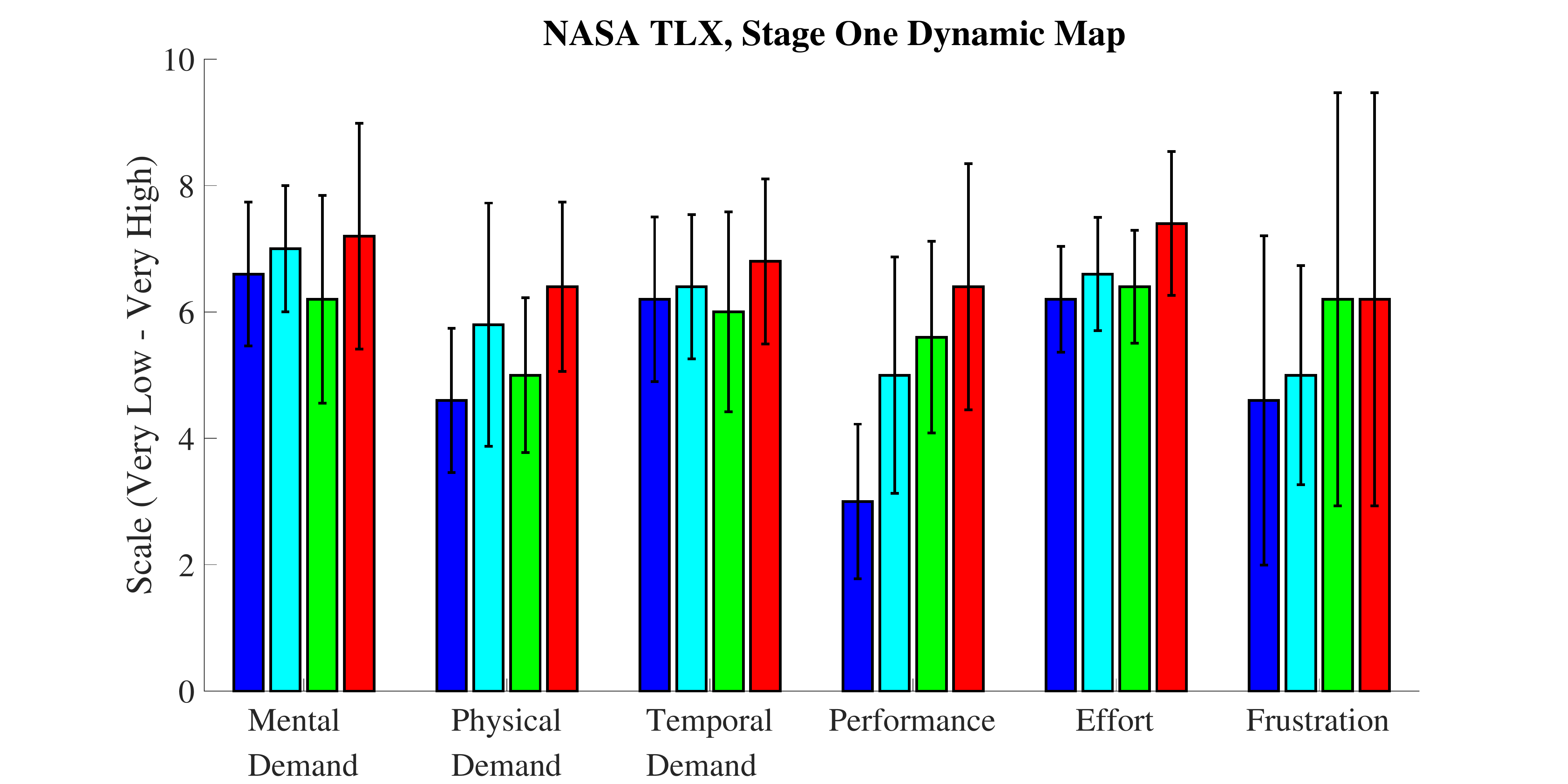}
 		\caption{}
        \end{subfigure}
 	\caption{\textbf{Each Subject's Results and NASA TLX Summary for Stage one Experiments}. Figures (a,b,d,e) indicate the mean and standard deviation of the five subjects' performance in the static and dynamic maps with five trials per method. The numerical values above the bars represent the mean values with the red texts representing the best performing cases. Figures (c,f) show the NASA TLX result average in the static and dynamic maps, respectively. (Blue: N-F, CYAN: F-H, GREEN: F-C, RED: Combo) }
 	\label{comparison_result_stage1}
\end{figure*}

\begin{table*}[!htp]
\centering
\caption{\textbf{Average and Individual Results for Each Methods in Experiment Stage One.} Mean and standard deviation are calculated based on the average performance of each participants (mean:\textbf{m}, standard deviation:\textbf{s}). Individual result represents the method with the highest number of good performers in the result of each subject.}
\label{t_benchmark_stage1}
\begin{tabular}{|cc|ccclclll|llllllll|}
\hline
\multicolumn{2}{|c|}{}                                                                      & \multicolumn{8}{c|}{\textbf{Static Map}}                                                                                                                                                                                                                                                                                                                                 & \multicolumn{8}{c|}{\textbf{Dynamic Map}}                                                                                                                                                                                                                                                                                                                                \\ \cline{3-18} 
\multicolumn{2}{|c|}{\multirow{-2}{*}{\textbf{}}}                                           & \multicolumn{2}{c|}{\textbf{\begin{tabular}[c]{@{}c@{}}C-T{[}s{]}\\ (m, s)\end{tabular}}} & \multicolumn{2}{c|}{\textbf{\begin{tabular}[c]{@{}c@{}}C-N\\ (m, s)\end{tabular}}} & \multicolumn{2}{c|}{\textbf{\begin{tabular}[c]{@{}c@{}}C-D{[}\%{]}\\ (m, s)\end{tabular}}} & \multicolumn{2}{c|}{\textbf{\begin{tabular}[c]{@{}c@{}}S-R{[}\%{]}\\ (m, s)\end{tabular}}} & \multicolumn{2}{c|}{\textbf{\begin{tabular}[c]{@{}c@{}}C-T{[}s{]}\\ (m, s)\end{tabular}}} & \multicolumn{2}{c|}{\textbf{\begin{tabular}[c]{@{}c@{}}C-N\\ (m, s)\end{tabular}}} & \multicolumn{2}{c|}{\textbf{\begin{tabular}[c]{@{}c@{}}C-D{[}\%{]}\\ (m, s)\end{tabular}}} & \multicolumn{2}{c|}{\textbf{\begin{tabular}[c]{@{}c@{}}S-R{[}\%{]}\\ (m, s)\end{tabular}}} \\ \hline
\multicolumn{1}{|c|}{}                                            & \textbf{N-F}            & \multicolumn{1}{c|}{72.79}                       & \multicolumn{1}{c|}{12.74}             & \multicolumn{1}{c|}{0.28}                   & \multicolumn{1}{l|}{0.10}            & \multicolumn{1}{c|}{0.91}                        & \multicolumn{1}{l|}{0.04}               & \multicolumn{1}{l|}{0.80}                                   & 0.14                         & \multicolumn{1}{l|}{\textbf{72.01}}              & \multicolumn{1}{l|}{8.56}              & \multicolumn{1}{l|}{\textbf{0.16}}           & \multicolumn{1}{l|}{0.16}           & \multicolumn{1}{l|}{0.89}                        & \multicolumn{1}{l|}{0.10}               & \multicolumn{1}{l|}{\textbf{0.80}}                          & 0.24                         \\ \cline{2-18} 
\multicolumn{1}{|c|}{}                                            & \textbf{W F-H}          & \multicolumn{1}{c|}{75.98}                       & \multicolumn{1}{c|}{17.01}             & \multicolumn{1}{c|}{0.32}                   & \multicolumn{1}{l|}{0.30}            & \multicolumn{1}{c|}{0.95}                        & \multicolumn{1}{l|}{0.04}               & \multicolumn{1}{l|}{0.80}                                   & 0.20                         & \multicolumn{1}{l|}{73.11}                       & \multicolumn{1}{l|}{8.10}              & \multicolumn{1}{l|}{0.20}                    & \multicolumn{1}{l|}{0.20}           & \multicolumn{1}{l|}{\textbf{0.91}}               & \multicolumn{1}{l|}{0.11}               & \multicolumn{1}{l|}{0.76}                                   & 0.26                         \\ \cline{2-18} 
\multicolumn{1}{|c|}{}                                            & \textbf{W F-C}          & \multicolumn{1}{c|}{\textbf{72.49}}              & \multicolumn{1}{c|}{11.46}             & \multicolumn{1}{c|}{\textbf{0.12}}          & \multicolumn{1}{l|}{0.17}          & \multicolumn{1}{c|}{\textbf{0.97}}               & \multicolumn{1}{l|}{0.04}               & \multicolumn{1}{l|}{\textbf{0.92}}                          & 0.17                         & \multicolumn{1}{l|}{74.60}                       & \multicolumn{1}{l|}{8.00}              & \multicolumn{1}{l|}{0.36}                    & \multicolumn{1}{l|}{0.26}           & \multicolumn{1}{l|}{0.88}                        & \multicolumn{1}{l|}{0.04}               & \multicolumn{1}{l|}{\textbf{0.80}}                          & 0.00                         \\ \cline{2-18} 
\multicolumn{1}{|c|}{\multirow{-4}{*}{\textbf{Method}}}           & \textbf{Combo}          & \multicolumn{1}{c|}{75.08}                       & \multicolumn{1}{c|}{12.83}             & \multicolumn{1}{c|}{0.32}                   & \multicolumn{1}{l|}{0.22}            & \multicolumn{1}{c|}{0.90}                        & \multicolumn{1}{l|}{0.14}               & \multicolumn{1}{l|}{0.80}                                   & 0.28                         & \multicolumn{1}{l|}{74.26}                       & \multicolumn{1}{l|}{7.89}              & \multicolumn{1}{l|}{0.32}                    & \multicolumn{1}{l|}{0.22}           & \multicolumn{1}{l|}{0.90}                        & \multicolumn{1}{l|}{0.06}               & \multicolumn{1}{l|}{0.76}                                   & 0.16                         \\ \hline
\multicolumn{2}{|c|}{\textbf{\begin{tabular}[c]{@{}c@{}}Individual \\ Result\end{tabular}}} & \multicolumn{2}{c|}{{\color[HTML]{000000} \textbf{N-F}}}                                  & \multicolumn{2}{c|}{{\color[HTML]{000000} \textbf{F-C}}}                           & \multicolumn{2}{c|}{\textbf{F-C}}                                                          & \multicolumn{2}{c|}{\textbf{F-C}}                                                          & \multicolumn{2}{c|}{\textbf{N-F, F-H}}                                                    & \multicolumn{2}{c|}{\textbf{N-F}}                                                  & \multicolumn{2}{c|}{\textbf{F-H}}                                                          & \multicolumn{2}{c|}{\textbf{N-F}}                                                          \\ \hline
\end{tabular}
\end{table*}

\subsection{Evaluation Procedure and Metrics}
All subjects had sufficient practice to acclimate themselves to the VR headset, the dynamics of SATYRR, the HMI, and the tasks prior to the experiments. Moreover, the gain $\mu$ from Eq. \ref{haptic_force_y}. was tuned based on each subject's preference for the haptic force feedback strength. Five trials per feedback case was conducted for all stage one maps and stage two static maps. Ten trials per feedback case was conducted for the stage two dynamic map considering its randomness and complexity. The order of the experiments for each method was randomized to remove the influence of the learning curve and fatigue in each experiment stage. To evaluate the performance of the cases carefully, we adopted the following evaluation metrics:\\   
\textbf{(1) Completion Time (C-T):} C-T represents the time measured when the operator finishes the task. A smaller number indicates a faster completion of the tasks. \\ 
\textbf{(2) Collision Number (C-N):} C-N represents the total number of collisions per case divided by the number of trials per case. A smaller number indicates fewer collisions with obstacles. This is considered as one of the most important metrics.\\
\textbf{(3) Completed Distance (C-D):} Since all maps have starting points and destinations in straight lines in the x-coordinate, C-D is the traversed distance divided by the distance from the start to the destination in the x-coordinate. A larger number indicates a higher completion percentage of the map.\\ 
\textbf{(4) Success Rate (S-R):} S-R is the number of successfully completed trials divided by the total number of trials. A number closer to one indicates a higher success rate. This is considered as one of the most important metrics. \\
\textbf{(5) NASA-TLX and Interviews:} Workload and user satisfaction were assessed through the NASA Task Load Index (TLX) \cite{hart1988development}. Interviews were conducted after all experiments to obtain qualitative feedback for each shared-control feedback case.

%% file: 4_results_discussion.tex
\section{RESULTS AND DISCUSSION}
\label{result}
\subsection{Quantitative Results from the Stage One Experiments}
\label{Results1}
% The experiment results for the static map are shown in figure #, where we see the feedback to controller case showed the best performance of all cases. Comparing to the no feedback base case, the controller case has a #\% decrease in completion time, #\% decrease in the number of collisions, and #\% increase in success rate. For the human feedback and combo cases, the performance differs insubstantially from the base case.

% In the dynamic map case shown in figure #, we can see the the base case with no feedback performed the best with the least completion time and the fewest number of collisions. The success rate for the feedback to controller case is slightly better than the base case, but the difference is insignificant.

% \begin{figure*}[t]
%      \centering
%  	\begin{subfigure}{0.24\linewidth}
%  		\includegraphics[width=\columnwidth]{figs/1__sub_each_completion time[s].pdf}
%  		\caption{}
%  	\end{subfigure} %\hspace{0.05cm}
%  	\begin{subfigure}{0.24\linewidth}
%  		\includegraphics[width=\columnwidth]{figs/1__sub_each_collision number.pdf}
%  		\caption{}
%  	\end{subfigure}
%  	\begin{subfigure}{0.24\linewidth}
%   		\includegraphics[width=\columnwidth]{figs/1__sub_each_distance.pdf}
%  		\caption{}
%  	\end{subfigure}
%  	\begin{subfigure}{0.24\linewidth}
%   		\includegraphics[width=\columnwidth]{figs/1__sub_each_success rate.pdf}
%  		\caption{}
%  	\end{subfigure}
% \end{figure*}

\subsubsection{Known Static Map} In all evaluation metrics from the aggravated results, F-C performed the best in the known static map (see Table. \ref{t_benchmark_stage1}). For the individual subjects' best cases, more than half of the subjects performed the best with F-C. Since the controller automatically corrected the subjects' turning mistakes with F-C, the subjects could focus on deciding the overall long-term path and rely on the robot to make small adjustments for a safer path. Nevertheless, the subjects did not know the robot's intention prior to the automatic turns and had to predict the compensated trajectory based on the obstacle locations from visual feedback. 

% In all evaluation metric, F-C outperformed other shared-control methods including the Combo Case (see Fig. \ref{comparison_result_stage1}). Furthermore, more than half of subjects show the best performance with the use of F-C. As the controller automatically corrects tight turning mistakes of users, the users put less effort to decide the general direction and let robot make adjustments to follow the safer path. Nevertheless, users cannot know what the robot is trying to do due to lack of prior indication, and this sometimes required a larger motion to cancel the automated command.  

As the second best method, N-F performed better than F-H and Combo overall since physical resistance and too much feedback in a known environment became a distraction. After the subjects learned the best paths and strategies to complete the tasks, the haptic feedback that physically resisted the subjects' desired motions induced more mental and physical effort that degraded the performance (See Fig. \ref{comparison_result_stage1}).

% We noticed that H-F became act as a distraction when all map and control interface were well-known since they already know the best strategy and behavior to accomplish the task. Especially, F-H adversely affects the completion time because the robot can be operated after the user overcomes the repulsive haptic force from HMI, and this leads the negative effect to Combo as well. However, we found that H-F has benefit to assist human to avoid the collision with obstacles based on results of each participants in S-R and C-N. This is because human can perceive danger via haptic force before the crash, bringing in modifying their movements to prevent collision. 

\subsubsection{Known Dynamic Map}
Interestingly, N-F showed the best performance in the dynamic map overall (see Table. \ref{t_benchmark_stage1}). For the individual subjects' best cases, N-F also outperformed other methods for more than half of the participants. Once the subjects got familiar with the map and discovered the optimal path, controlling the robot's forward motion and timing the moving obstacles became the key challenge. To overcome this, the subjects needed full control the robot to perform more dynamic and responsive motions, so any additional haptic feedback or controller compensation became disturbance and added uncertainty. Moreover, the visual feedback was sufficient for predicting the obstacles' trajectories since the obstacle locations and velocities were known to the subject, so any additional feedback that require mental effort to process was detrimental to the performance.

\subsection{Qualitative Results from the Stage One Experiments}
In both the static and dynamic maps, most subjects preferred N-F, reporting the least frustration, least physical demand, and best perceived performance through the NASA TLX (see Fig. \ref{comparison_result_stage1}). F-C showed slightly less mental demand compared to N-F since the subjects could rely on the robot to make small adjustments for obstacle avoidance. However, F-C sometimes generated unexpected robot trajectory especially when the obstacles were moving, resulting in more frustration and worse perceived performance. With Combo, the subjects had to put in more mental and physical effort since the haptic force and the compensated controller became disturbance and uncertainty to the control in a known environment, which led to the most effort and frustration with the worst perceived performance.

\subsection{Quantitative Results from the Stage Two Experiments}
\label{quan_result_stage2}

\subsubsection{Unknown Static Dark Map}
The advantages of using force feedback are revealed in unfamiliar environments. In the unknown static dark map, Combo outperformed the other methods in all evaluation metrics (see Table \ref{t_table_avg_stage2}) and likewise in the results for each participant as shown in Fig. \ref{comparison_result_stage2}. Due to the reduced visual feedback and the randomized paths, the operators had to learn and rely more on the compensated controller and the haptic feedback combination. This trend is further demonstrated with F-H and F-C where both feedback cases showed better performance than N-F, suggesting that more feedback is consistently more useful in unfamiliar environments with reduced visual feedback. The subjects could intuitively understand the robot's automatic turning intention since the activation distance for F-H in Combo is further than F-C as described in the experiment section. The subjects also had the final decision power over the automatic turns by fighting against F-H that reflected F-C.

\subsubsection{Unknown Static Bright Map}
\begin{figure*}[t]
     \centering
 	\begin{subfigure}{0.44\linewidth}
 		\includegraphics[width=\columnwidth]{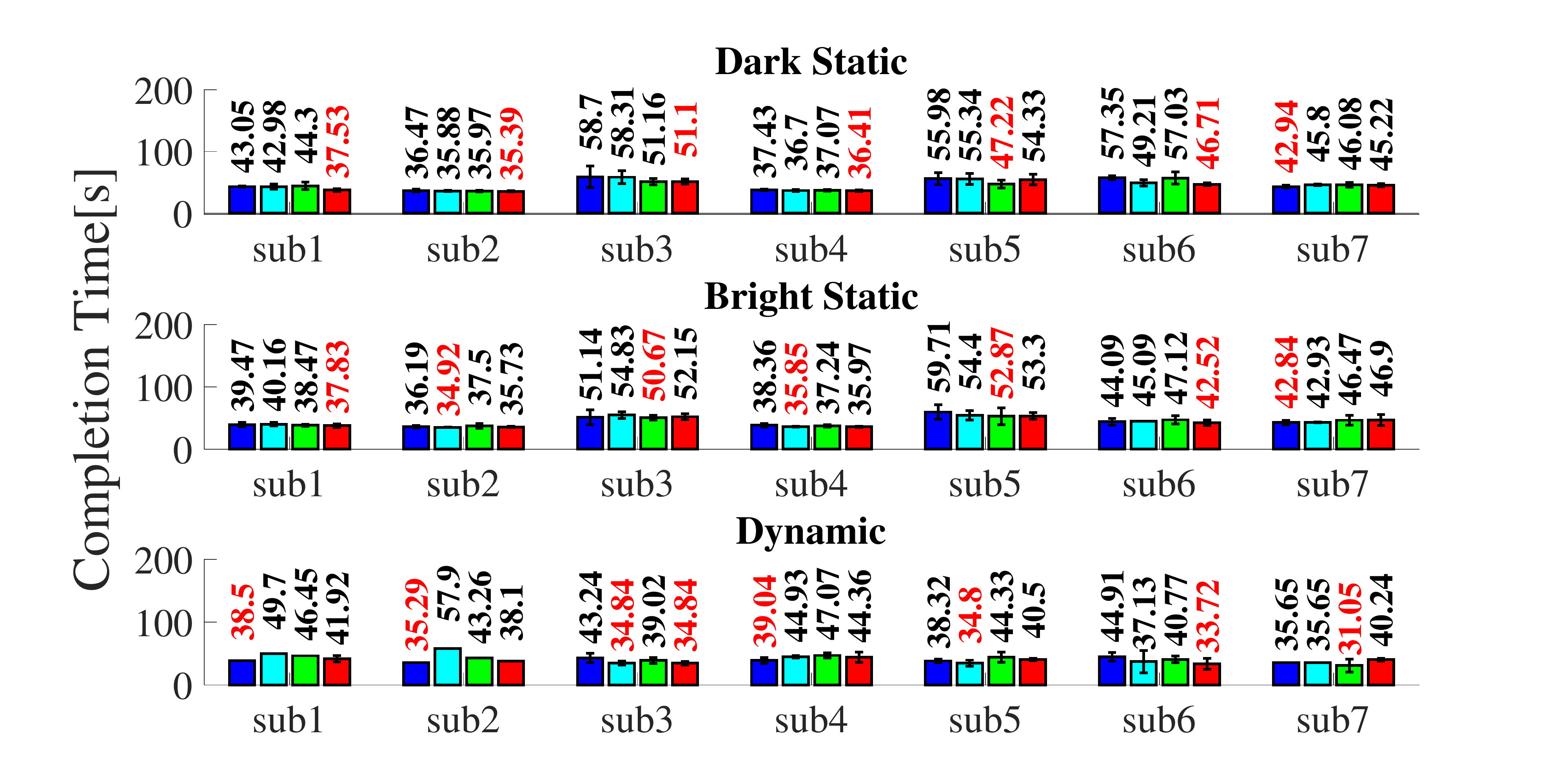}
 		\caption{}
 	\end{subfigure} %\hspace{0.05cm}
 	\begin{subfigure}{0.44\linewidth}
 		\includegraphics[width=\columnwidth]{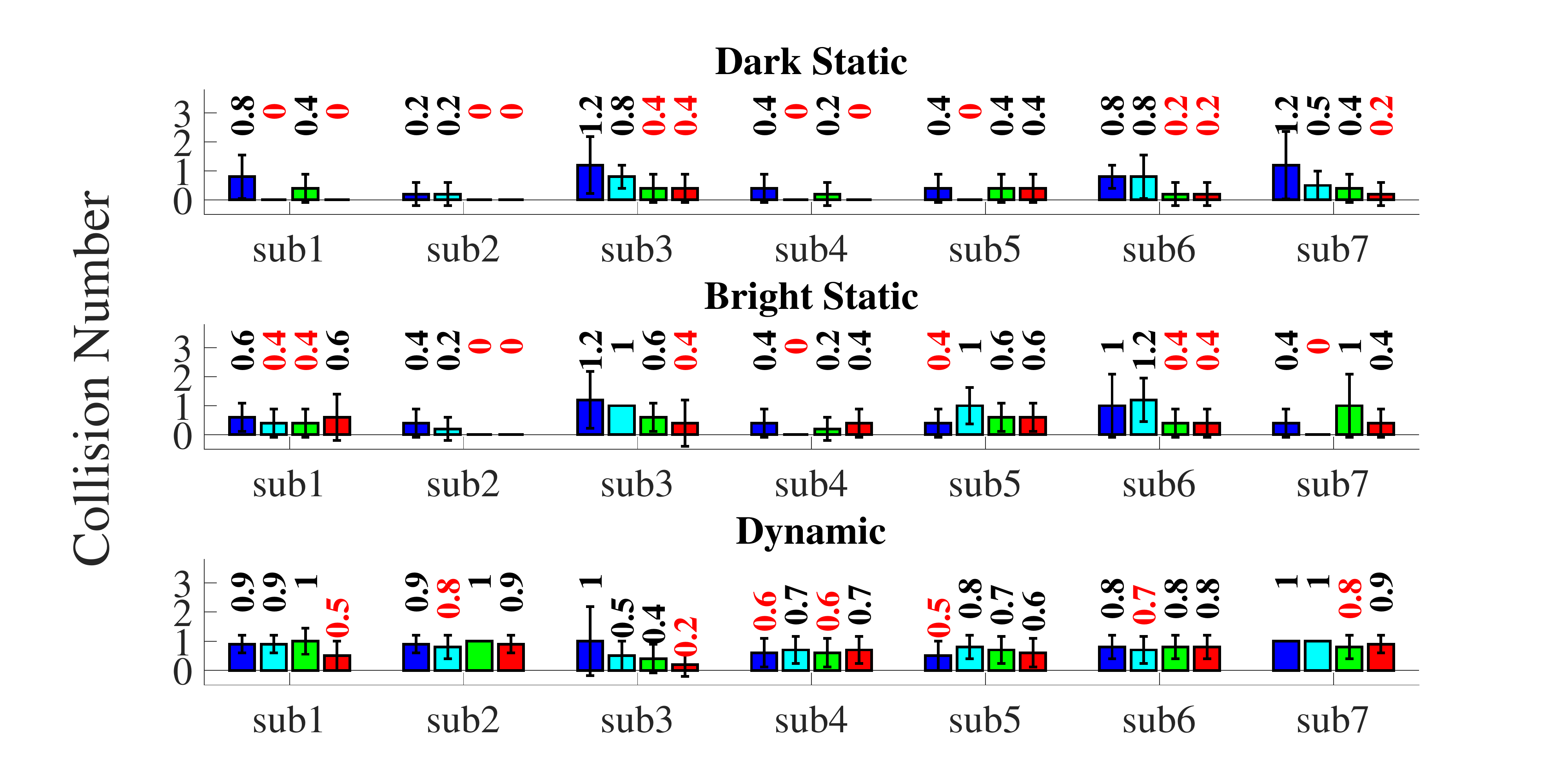}
 		\caption{}
 	\end{subfigure} %\hspace{0.05cm}

 	\begin{subfigure}{0.44\linewidth}
  		\includegraphics[width=\columnwidth]{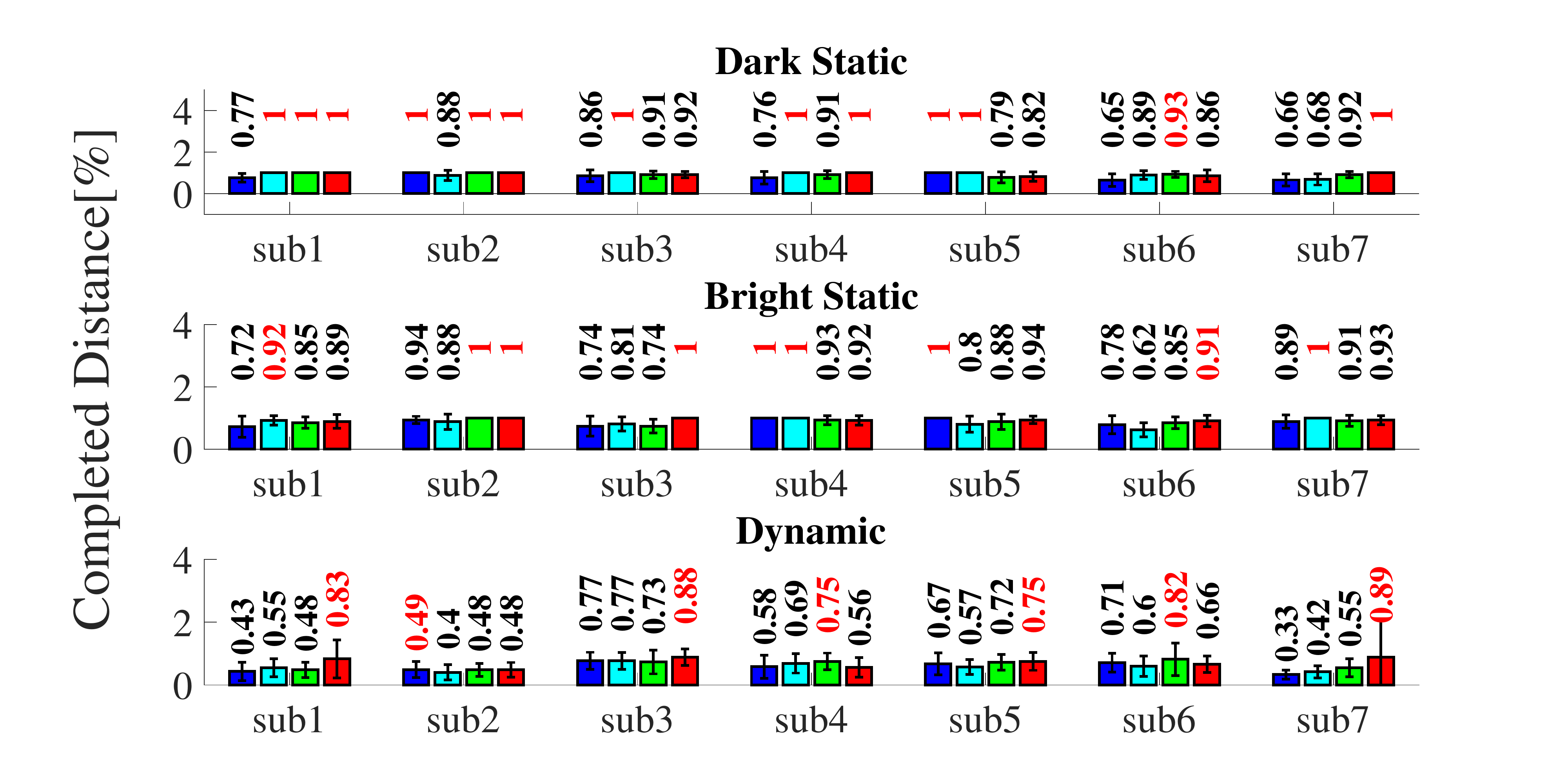}
 		\caption{}
 	\end{subfigure}
 	\begin{subfigure}{0.44\linewidth}
  		\includegraphics[width=\columnwidth]{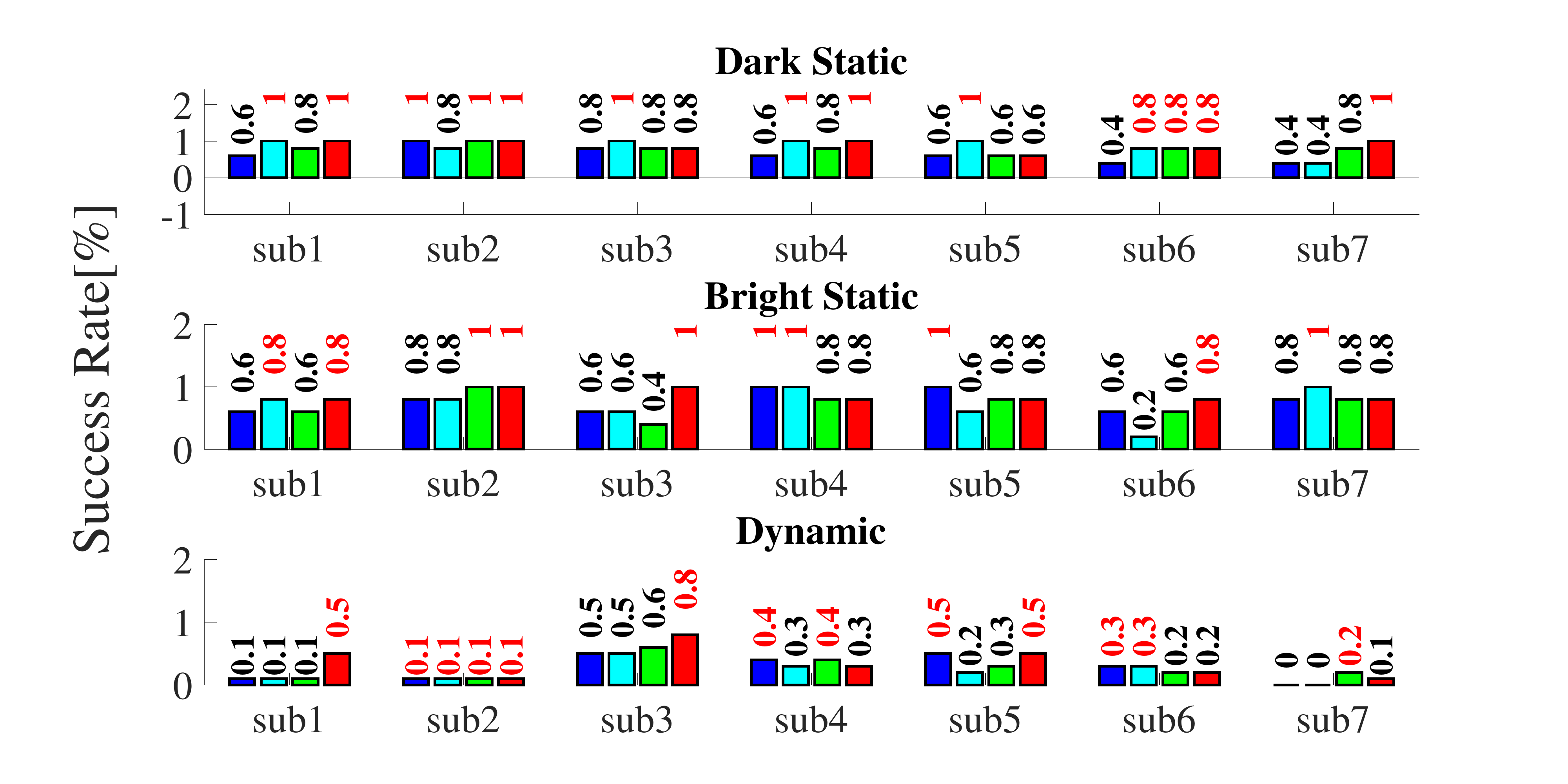}
 		\caption{}
 	\end{subfigure}

        \begin{subfigure}{0.325\linewidth}
  		\includegraphics[width=\columnwidth]{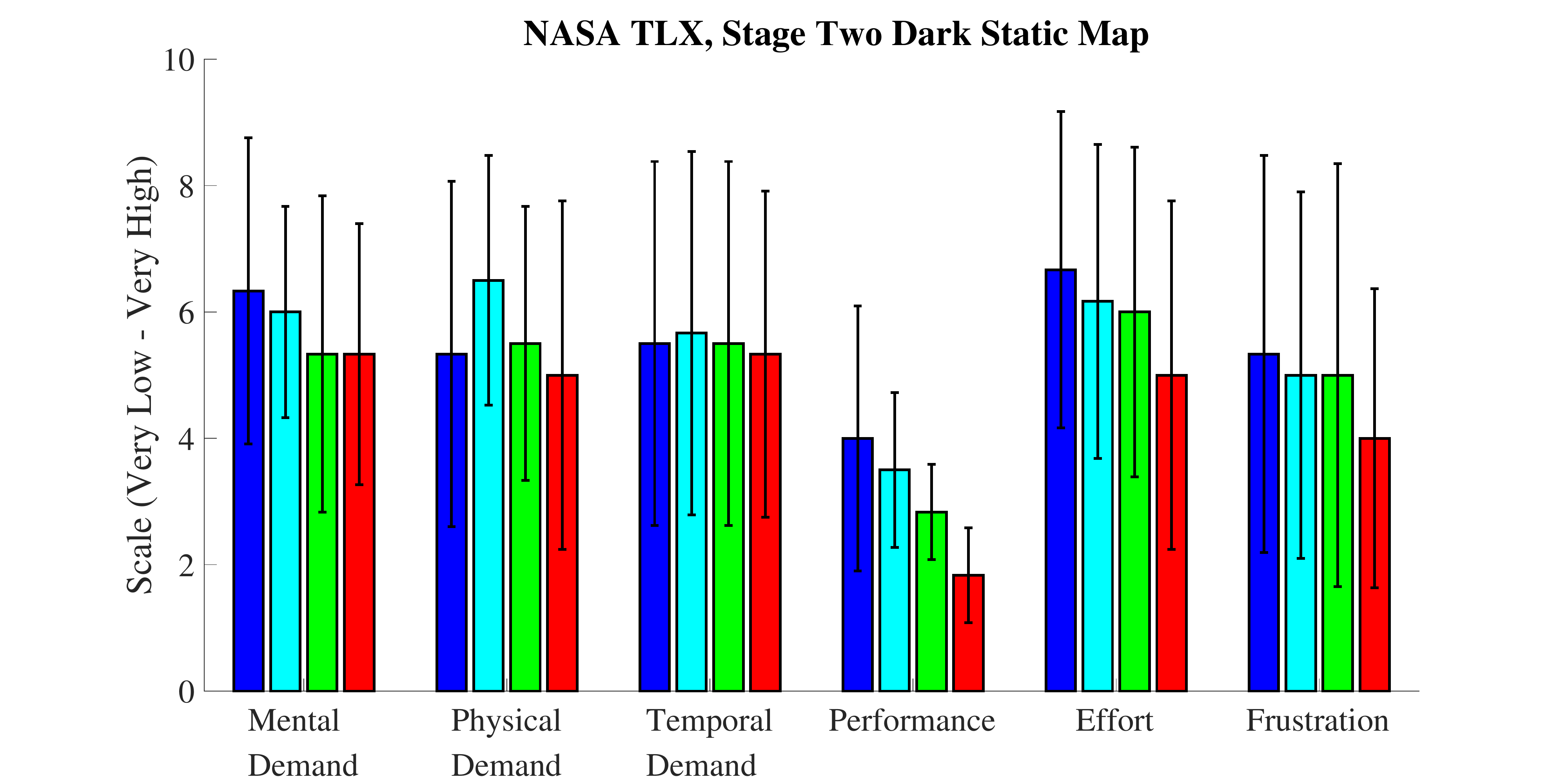}
 		\caption{}
 	\end{subfigure}
 	\begin{subfigure}{0.325\linewidth}
  		\includegraphics[width=\columnwidth]{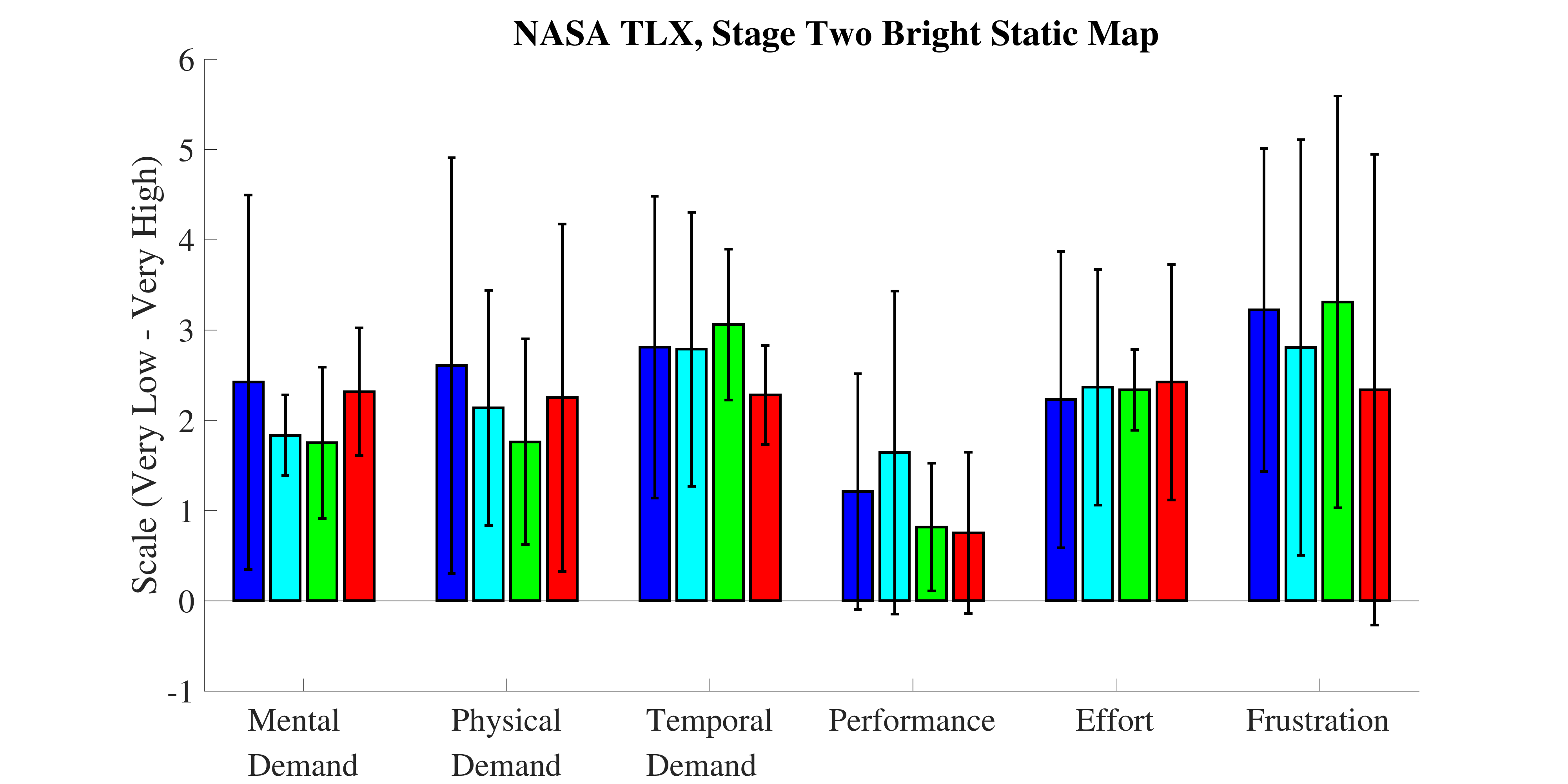}
 		\caption{}
 	\end{subfigure}
        \begin{subfigure}{0.325\linewidth}
  		\includegraphics[width=\columnwidth]{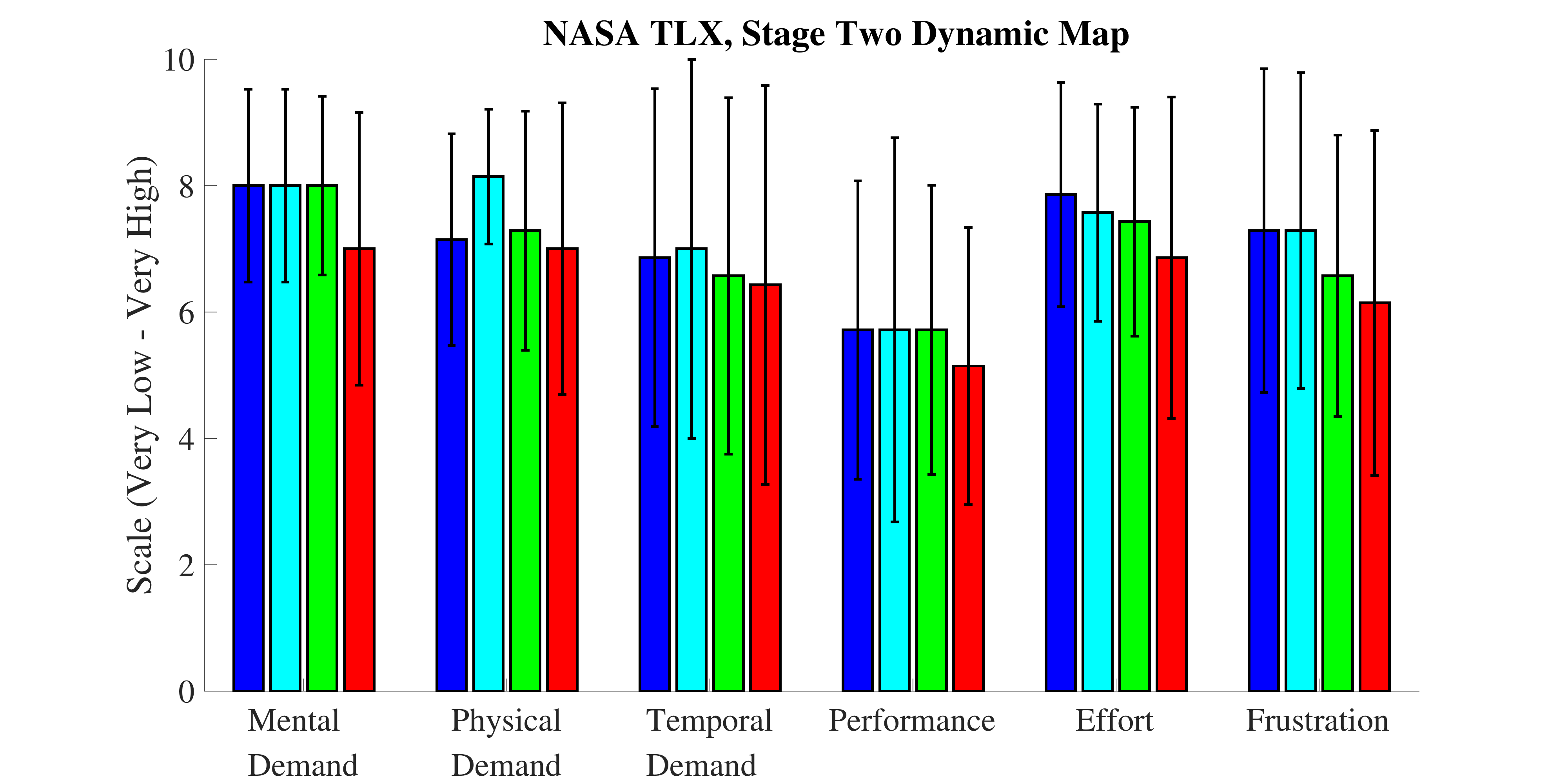}
 		\caption{}
 	\end{subfigure}
 	\caption{\textbf{Result of Each Subjects in Experiment Stage Two}. Figures (a-d) represent the mean and standard deviation of each subject's static maps (bright and dark) and dynamic map performance for the four feedback methods. The numerical values above the bars represent the mean values with the red texts representing the best performing cases. Figures (e-g) show the NASA TLX result in the static maps (bright,dark) and dynamic map. (Blue: N-F, CYAN: F-H, GREEN: F-C, RED: Combo)}
 	\label{comparison_result_stage2}
\end{figure*}

In the unknown static bright map, Combo also showed the best performance in both the aggregated results and the individual outcomes (see Table \ref{t_table_avg_stage2} and Fig. \ref{comparison_result_stage2}). Both F-H and F-C showed marginally better average performance in most evaluation metrics but slightly lower S-R than N-F. All feedback cases showed less C-N compared to N-F, indicating that the feedback is helpful for avoiding collisions in the unknown static bright map. Note that the C-N is higher in the bright static map than the dark static map since the subjects were more cautious and moved slower in the dark map as shown through C-T.

\subsubsection{Unknown Dynamic Map}
In this complex environment where numerous obstacles moved with random velocities, Combo showed the best performance across almost all metrics (see Table \ref{t_table_avg_stage2}). This indicates that Combo is advantageous in complicated and visual sensory overloading environments where the operator has to make many predictions and decisions. Combo outperformed F-H and F-C since the subjects could rely on F-C for automatic turns while using F-H to understand the robot's intention quickly. The automatic turns in F-C and Combo protected the subjects from many close collision encounters, demonstrated with less C-N and higher C-D. However, N-F surpassed Combo in C-T for the individual subjects' best cases since the subjects' decisions for the longer distance path sometimes conflicted with the repulsive force feedback for the shorter distance path. The C-T of F-H and F-C further supports that any instance of such inconsistency could slow down the commanded speed from the subjects in this complex and dynamic map. Moreover, F-H showed the worst S-R since interpreting the additional force feedback without the robot making automatic turns simultaneously required extra mental effort that the subjects could not afford when there were too many obstacles to keep track of. 

\label{Results3}

\begin{table*}[!htp]
\centering
\caption{\textbf{Average and Individual Results for Each Methods in Experiment Stage Two.} Mean and standard deviation are calculated based on the average performance of each participants (mean:\textbf{m}, standard deviation:\textbf{s}). Individual result represents the method with the highest number of good performers in the result of each subject.}
\label{t_table_avg_stage2}
\setlength{\tabcolsep}{2pt}
\begin{tabular}{|cc|cccccccc|cccccccc|cccccccc|}
\hline
\multicolumn{2}{|c|}{\multirow{2}{*}{\textbf{}}}                                                     & \multicolumn{8}{c|}{\textbf{Static Dark Map}}                                                                                                                                                                                                                                                                                                                            & \multicolumn{8}{c|}{\textbf{Static Bright Map}}                                                                                                                                                                                                                                                                                                                             & \multicolumn{8}{c|}{\textbf{Dynamic Map}}                                                                                                                                                                                                                                                                                                                                  \\ \cline{3-26} 
\multicolumn{2}{|c|}{}                                                                               & \multicolumn{2}{c|}{\textbf{\begin{tabular}[c]{@{}c@{}}C-T{[}s{]}\\ (m, s)\end{tabular}}} & \multicolumn{2}{c|}{\textbf{\begin{tabular}[c]{@{}c@{}}C-N\\ (m, s)\end{tabular}}} & \multicolumn{2}{c|}{\textbf{\begin{tabular}[c]{@{}c@{}}C-D{[}\%{]}\\ (m, s)\end{tabular}}} & \multicolumn{2}{c|}{\textbf{\begin{tabular}[c]{@{}c@{}}S-R{[}\%{]}\\ (m, s)\end{tabular}}} & \multicolumn{2}{c|}{\textbf{\begin{tabular}[c]{@{}c@{}}C-T{[}s{]}\\ (m, s)\end{tabular}}} & \multicolumn{2}{c|}{\textbf{\begin{tabular}[c]{@{}c@{}}C-N\\ (m, s)\end{tabular}}}    & \multicolumn{2}{c|}{\textbf{\begin{tabular}[c]{@{}c@{}}C-D{[}\%{]}\\ (m, s)\end{tabular}}} & \multicolumn{2}{c|}{\textbf{\begin{tabular}[c]{@{}c@{}}S-R{[}\%{]}\\ (m, s)\end{tabular}}} & \multicolumn{2}{c|}{\textbf{\begin{tabular}[c]{@{}c@{}}C-T{[}s{]}\\ (m, s)\end{tabular}}} & \multicolumn{2}{c|}{\textbf{\begin{tabular}[c]{@{}c@{}}C-N\\ (m, s)\end{tabular}}}   & \multicolumn{2}{c|}{\textbf{\begin{tabular}[c]{@{}c@{}}C-D{[}\%{]}\\ (m, s)\end{tabular}}} & \multicolumn{2}{c|}{\textbf{\begin{tabular}[c]{@{}c@{}}S-R{[}\%{]}\\ (m, s)\end{tabular}}} \\ \hline
\multicolumn{1}{|c|}{\multirow{4}{*}{\textbf{Method}}}                & \textbf{N-F}                 & \multicolumn{1}{c|}{47.41}                       & \multicolumn{1}{c|}{9.64}              & \multicolumn{1}{c|}{0.71}                    & \multicolumn{1}{c|}{0.39}           & \multicolumn{1}{c|}{0.81}                        & \multicolumn{1}{c|}{0.14}               & \multicolumn{1}{c|}{0.62}                                   & 0.21                         & \multicolumn{1}{c|}{44.54}                       & \multicolumn{1}{c|}{8.27}              & \multicolumn{1}{c|}{0.62}                      & \multicolumn{1}{c|}{0.33}            & \multicolumn{1}{c|}{0.86}                        & \multicolumn{1}{c|}{0.12}               & \multicolumn{1}{c|}{0.77}                                   & 0.17                         & \multicolumn{1}{c|}{39.27}                       & \multicolumn{1}{c|}{3.60}              & \multicolumn{1}{c|}{0.81}                     & \multicolumn{1}{c|}{0.19}            & \multicolumn{1}{c|}{0.56}                        & \multicolumn{1}{c|}{0.16}               & \multicolumn{1}{c|}{0.27}                                   & 0.20                         \\ \cline{2-26} 
\multicolumn{1}{|c|}{}                                                & \textbf{F-H}                 & \multicolumn{1}{c|}{46.31}                       & \multicolumn{1}{c|}{8.62}              & \multicolumn{1}{c|}{0.32}                    & \multicolumn{1}{c|}{0.36}           & \multicolumn{1}{c|}{0.92}                        & \multicolumn{1}{c|}{0.11}               & \multicolumn{1}{c|}{0.85}                                   & 0.22                         & \multicolumn{1}{c|}{44.02}                       & \multicolumn{1}{c|}{8.07}              & \multicolumn{1}{c|}{0.54}                      & \multicolumn{1}{c|}{0.51}            & \multicolumn{1}{c|}{0.86}                        & \multicolumn{1}{c|}{0.13}               & \multicolumn{1}{c|}{0.71}                                   & 0.27                         & \multicolumn{1}{c|}{42.13}                       & \multicolumn{1}{c|}{9.01}              & \multicolumn{1}{c|}{0.77}                     & \multicolumn{1}{c|}{0.16}            & \multicolumn{1}{c|}{0.57}                        & \multicolumn{1}{c|}{0.13}               & \multicolumn{1}{c|}{0.21}                                   & 0.16                         \\ \cline{2-26} 
\multicolumn{1}{|c|}{}                                                & \textbf{F-C}                 & \multicolumn{1}{c|}{45.54}                       & \multicolumn{1}{c|}{7.43}              & \multicolumn{1}{c|}{0.28}                    & \multicolumn{1}{c|}{0.15}           & \multicolumn{1}{c|}{0.92}                        & \multicolumn{1}{c|}{0.07}               & \multicolumn{1}{c|}{0.80}                                   & 0.11                         & \multicolumn{1}{c|}{44.33}                       & \multicolumn{1}{c|}{6.54}              & \multicolumn{1}{c|}{0.45}                      & \multicolumn{1}{c|}{0.32}            & \multicolumn{1}{c|}{0.87}                        & \multicolumn{1}{c|}{0.08}               & \multicolumn{1}{c|}{0.71}                                   & 0.19                         & \multicolumn{1}{c|}{41.70}                       & \multicolumn{1}{c|}{5.51}              & \multicolumn{1}{c|}{0.75}                     & \multicolumn{1}{c|}{0.21}            & \multicolumn{1}{c|}{0.64}                        & \multicolumn{1}{c|}{0.13}               & \multicolumn{1}{c|}{0.27}                                   & 0.17                         \\ \cline{2-26} 
\multicolumn{1}{|c|}{}                                                & \textbf{Combo}               & \multicolumn{1}{c|}{\textbf{43.81}}              & \multicolumn{1}{c|}{7.51}              & \multicolumn{1}{c|}{\textbf{0.17}}           & \multicolumn{1}{c|}{0.17}           & \multicolumn{1}{c|}{\textbf{0.94}}               & \multicolumn{1}{c|}{0.07}               & \multicolumn{1}{c|}{\textbf{0.88}}                          & 0.15                         & \multicolumn{1}{c|}{\textbf{43.48}}              & \multicolumn{1}{c|}{7.44}              & \multicolumn{1}{c|}{\textbf{0.40}}             & \multicolumn{1}{c|}{0.20}            & \multicolumn{1}{c|}{\textbf{0.94}}               & \multicolumn{1}{c|}{0.04}               & \multicolumn{1}{c|}{\textbf{0.85}}                          & 0.09                         & \multicolumn{1}{c|}{\textbf{39.09}}              & \multicolumn{1}{c|}{3.80}              & \multicolumn{1}{c|}{\textbf{0.65}}            & \multicolumn{1}{c|}{0.25}            & \multicolumn{1}{c|}{\textbf{0.72}}               & \multicolumn{1}{c|}{0.16}               & \multicolumn{1}{c|}{\textbf{0.35}}                          & 0.25                         \\ \hline
\multicolumn{2}{|c|}{\textbf{\begin{tabular}[c]{@{}c@{}}Individual\\ Result\end{tabular}}} & \multicolumn{2}{c|}{\textbf{CB}}                                                          & \multicolumn{2}{c|}{\textbf{CB}}                                                   & \multicolumn{2}{c|}{\textbf{CB}}                                                           & \multicolumn{2}{c|}{\textbf{\begin{tabular}[c]{@{}c@{}}F-H,\\ CB\end{tabular}}}            & \multicolumn{2}{c|}{\textbf{\begin{tabular}[c]{@{}c@{}}F-H, F-C,\\ CB\end{tabular}}}      & \multicolumn{2}{c|}{\textbf{\begin{tabular}[c]{@{}c@{}}F-H, F-C, \\ CB\end{tabular}}} & \multicolumn{2}{c|}{\textbf{\begin{tabular}[c]{@{}c@{}}F-H,\\ CB\end{tabular}}}            & \multicolumn{2}{c|}{\textbf{CB}}                                                           & \multicolumn{2}{c|}{\textbf{N-F}}                                                         & \multicolumn{2}{c|}{\textbf{\begin{tabular}[c]{@{}c@{}}F-H, F-C,\\ CB\end{tabular}}} & \multicolumn{2}{c|}{\textbf{CB}}                                                           & \multicolumn{2}{c|}{\textbf{\begin{tabular}[c]{@{}c@{}}N-F,\\ CB\end{tabular}}}            \\ \hline
\end{tabular}
\end{table*}

\subsection{Qualitative Results from the Stage Two Experiments}
\label{qual_result_stage2}
The NASA TLX results shown in Fig. \ref{comparison_result_stage2} denote that the subjects preferred the feedback cases over N-F in all stage two unknown maps. In our interviews, the majority of the subjects also indicated their preference for Combo over other feedback cases. In the dark static and dynamic maps, Combo followed by F-C and F-H showed the least mental, physical, and temporal demands while requiring the least effort and generating less frustration. For path planning, most subjects indicated the use of the visual feedback in the longer distance and the force feedback in the shorter distance. F-H by itself was helpful for warning dangerous encounters with obstacles, but avoiding obstacles without the robot's automatic turns was physically demanding. Likewise, when the subjects were using only F-C, the robot corrected most subjects' mistakes effectively but the automatic turns sometimes overcompensated or occurred unexpectedly, requiring more physical effort to teleoperate the robot.

\subsection{Summary and Suggestions}
The key takeaways from both stages of experiments and suggestions for future implementations are listed below: \\
(1) Force feedback from the obstacles can degrade performance when the environments are known to the operator unless the maps are static and no physical haptic force is provided. However, the force feedback is consistently helpful and preferred when the maps are unfamiliar. \\
(2) When the operator is overloaded by or lacks visual feedback from unfamiliar environments, the force feedback shows more significant boost in performance.\\
(3) Humans can intuitively use force feedback for reactive short-distance planning and visual feedback for long-distance planning. This strategy is important for avoiding dynamic obstacles.\\
(4) In unfamiliar maps, conveying the turning intention of the robot to the operator boosts performance and is preferred over other feedback cases as demonstrated by Combo's results.

\subsection{Limitations}
\label{Limitations}
Although force feedback in our shared-control framework showed better performance in various conditions, our research has several limitations to be further improved. First, a few subjects who felt nauseous from VR were unable to start the experiments. This could be compensated by replacing VR with a 2D screen that provides less visual information. Second, the operators controlled and received force feedback from the robot yaw through the less intuitive frontal plane motion due to the HMI configurations. Matching the robot's yaw with the human's yaw motion could further improve the performance of the force feedback methods. Third, the experiments were conducted in simulations that assumed little time delay in communication and known obstacle locations. More communication delay would likely lead to more collisions, but shared-control methods have shown to reduce the number of collisions over direct tele-operation with delay \cite{storms2017shared}. Both time delay and obstacle location estimation on the physical hardware will be explored in future work.

% (1) Humans tend to rely on bilateral feedback in shared-control for humanoid robot telelocomotion if the map unfamiliar.\\
% (2) Bilateral feedback is not useful when the map and tasks are well-known and consistent. \\
% (3) The F-C that automatically adjusts the motion of the robot shows the benefit in most cases, but this is disturbed if the F-C operates against human intention.\\
% (4) The F-H has the benefit for human to notice the danger in advance, but this is usually better to use with F-C together because this requires less physical demanding.

\label{Summary}

%% file: 5_conclusion.tex
\section{CONCLUSION}
\label{concl}
In this study, we explore various shared-control methods with force feedback through whole-body telelocomotion on a humanoid robot for obstacle avoidance in diverse environments. A Time-Derivative Sigmoid Function (TDSF) is proposed to generate a more intuitive force feedback. Two stages of extensive experiments encompassing five different maps were conducted using a humanoid robot SATYRR, an HMI, and a VR headset. The experiment results indicate that when the environments are familiar to the operator, force feedback often becomes a disturbance and introduces uncertainty to the control. However, force feedback improves performance and is preferred when the environments are unfamiliar to the operator. Moreover, when the operator lacks visual information or has visual sensory overload due to complicated environments, providing force feedback further boosts performance and improves user experience. In summary, our work evaluates the strengths and weaknesses of different force feedback shared-control methods under various obstacle environments for whole-body telelocomotion of a humanoid robot.